\documentclass[11pt,a4paper]{article}

\usepackage[utf8]{inputenc}
\usepackage[T1]{fontenc}
\usepackage{mathptmx}  
\usepackage[margin=1in]{geometry}
\usepackage{amsmath,amssymb,amsfonts}
\usepackage{graphicx}
\usepackage{booktabs}
\usepackage{hyperref}
\usepackage[ruled,vlined,linesnumbered]{algorithm2e}
\usepackage{multirow}
\usepackage{xcolor}
\usepackage{float}
\usepackage{enumitem}
\usepackage{caption}
\usepackage{url}

\hypersetup{colorlinks=true,linkcolor=blue,citecolor=blue,urlcolor=blue}
\SetAlgoLined
\SetKwComment{Comment}{// }{}
\title{\textbf{Brainstacks: Cross-Domain Cognitive Capabilities\\via Frozen MoE-LoRA Stacks for Continual LLM Learning}}
\author{
Mohammad R. Abu Ayyash\\
Brains Build Research, Ramallah, Palestine\\
\texttt{mohammadrabuayyash@gmail.com}
}
\date{April 2026}

\begin{document}
\maketitle

\begin{figure}[H]
\centering
\includegraphics[width=\textwidth]{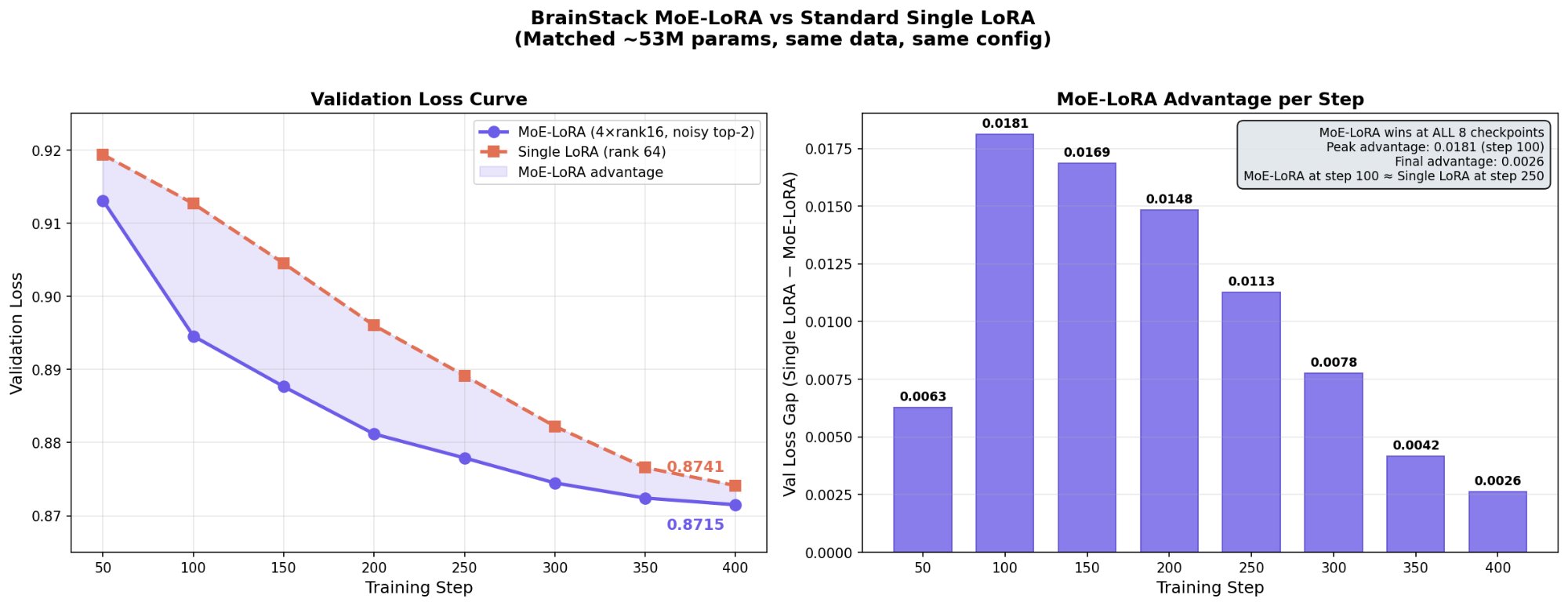}
\end{figure}
\vspace{-1em}

\begin{abstract}
We present Brainstacks, a novel modular architecture for continual multi-domain fine-tuning of large language models that packages domain expertise as frozen adapter stacks which compose additively on a shared frozen base model at inference. Brainstacks is built on five interlocking components: (1)~a Mixture-of-Experts LoRA (MoE-LoRA) building block with Shazeer-style noisy top-2 routing across all seven transformer projection matrices under QLoRA 4-bit quantization with rsLoRA scaling; (2)~an inner loop that performs residual boosting by freezing trained MoE-LoRA stacks and adding new ones to learn what previous stacks left uncaptured; (3)~an outer loop that trains sequential domain-specific stacks for continual learning with curriculum-ordered domain dependencies; (4)~null-space projection via randomized SVD that constrains new stacks to write in subspaces orthogonal to previously claimed directions, achieving zero forgetting when domains are evaluated in isolation; (5)~an outcome-based sigmoid meta-router trained on empirically discovered domain-combination targets that selectively weights stacks at inference, enabling true cross-domain composition. We additionally conduct two boundary experiments: (6)~a Partitioned Subspace Network (PSN) pretraining experiment testing the architecture on a randomly initialized model; and (7)~a per-domain reinforcement learning (DPO/GRPO) experiment validating that the stacking primitive is compatible with post-SFT alignment. We validate Brainstacks on TinyLlama-1.1B (4 domains, 9 stacks) and Gemma 3 12B IT (5 domains, 10 stacks), demonstrating that MoE-LoRA achieves $2.5\times$ faster convergence than parameter-matched single LoRA, that residual boosting breaks through the single-stack ceiling, and that the full routed system recovers generation quality destroyed by ungated stack accumulation. The central finding reframes what fine-tuning does: the outcome-based router discovers that domain stacks encode transferable cognitive primitives---instruction-following clarity, numerical reasoning, procedural logic, chain-of-thought structure---rather than domain-specific knowledge, with medical prompts optimally routing to chat+math stacks in 97\% of cases despite zero medical data in those stacks. Fine-tuning injects composable capabilities that transfer across domain boundaries, not knowledge retrieval.
\end{abstract}

\noindent\textbf{Code:} \url{https://github.com/achelousace/brainstacks}

\section{Introduction}

Current approaches to extending the capabilities of large language models (LLMs) remain fundamentally monolithic. Large language models are typically fine-tuned on mixed datasets in a single training run, coupling all domain knowledge into shared parameters. This creates three fundamental problems: (1)~adding a new domain requires retraining from scratch or risking catastrophic forgetting; (2)~there is no mechanism to remove or update individual domain capabilities post-deployment; and (3)~inference applies all learned knowledge uniformly regardless of the input, with no ability to selectively activate relevant expertise. When an organization needs a model that handles both medical question answering and code generation, the standard practice is either full fine-tuning on a combined dataset or sequential fine-tuning that inevitably suffers from catastrophic forgetting [McCloskey and Cohen, 1989, Kirkpatrick et al., 2017].

Parameter-efficient fine-tuning methods like LoRA (Hu et al., 2022) reduce training cost but do not address modularity. Mixture-of-Experts extensions (MoLoRA, MixLoRA, LoRAMoE) introduce conditional computation within a single training phase but lack continual learning capability. Continual learning methods (EWC, PackNet) protect prior knowledge through regularization but remain monolithic, cannot compose domain capabilities at inference, and provide no architectural guarantee against forgetting.

Brainstacks addresses all three problems through a single architectural primitive: the frozen MoE-LoRA stack. Each domain is trained as a set of stacked residual MoE-LoRA adapters, then permanently frozen. New domains train on top of frozen stacks with gradient constraints that project updates into the null space of prior domain activations. At inference, a meta-router selectively activates relevant domain stacks per prompt. The result is a modular system where domains can be independently added, removed, or updated without affecting other capabilities.

Our contributions are:

\begin{enumerate}[leftmargin=*]
\item A novel two-loop training architecture inspired by mechanisms of human brain function, combining residual boosting (inner loop) with continual domain stacking (outer loop) through the same MoE-LoRA primitive. Each new stack trains on the same loss but over a shifted landscape---frozen stacks have already corrected part of the output, so the active stack naturally learns what remains uncaptured, deepening domain capability through iterative refinement. Applied to all 7 transformer projections with Shazeer-style noisy routing and rsLoRA scaling.

\item Null-space gradient projection using randomized SVD that enforces orthogonality between domain subspaces, providing zero forgetting when domains are evaluated in isolation. The meta-router completes the anti-forgetting system by selectively gating stacks at inference.

\item An outcome-based sigmoid meta-router that discovers optimal domain combinations through exhaustive loss measurement, replacing label-based routing with empirical capability composition.

\item The central empirical finding that domain stacks learn transferable cognitive primitives (instruction-following clarity, numerical reasoning, procedural logic, chain-of-thought structure) rather than only domain-specific knowledge, with the outcome-based router providing the first direct evidence of this phenomenon: medical prompts route to chat+math stacks 97\% of the time, with zero medical data in those stacks.

\item A disk-offloaded inference system enabling selective stack loading per prompt, allowing arbitrarily many domain stacks with constant GPU memory, which we term the Superposition LLM principle.
\end{enumerate}

\section{Related Work}

\subsection{MoE-LoRA Methods}

Several works combine Mixture-of-Experts with LoRA adapters. MoLoRA (Zadouri et al., 2023) uses full soft routing over all experts but applies MoE only to some layers and uses no load-balance loss. MixLoRA (Li et al., 2024) applies top-K routing to FFN layers only, with a single shared LoRA on attention projections. LoRAMoE (Dou et al., 2024) addresses world-knowledge forgetting during instruction tuning through a localized balancing constraint that splits experts into knowledge-preservation and downstream-task groups. SiRA (Zhu et al., 2023) uses top-K with capacity limits and expert dropout. HydraLoRA (Tian et al., 2024) shares the A matrix across experts with per-expert B matrices.

Brainstacks differs from all prior MoE-LoRA work in three independently novel aspects verified against the literature: (1)~Shazeer-style learnable noise injection in the router via a dedicated \texttt{noise\_linear} layer with softplus activation, never previously applied to LoRA experts; (2)~MoE routing on all 7 transformer projections including attention (q, k, v, o), whereas prior work limits MoE to FFN or uses single LoRA on attention; and (3)~rsLoRA rank-stabilized scaling ($\alpha/\sqrt{r}$) combined with MoE-LoRA, which no prior work employs. Furthermore, no existing work freezes and residually stacks MoE-LoRA modules for continual learning; prior adapter-stacking methods stack standard single LoRA, not routed expert ensembles.

\subsection{Continual Learning with Adapters}

Progressive neural networks (Rusu et al., 2016) introduce lateral connections between frozen task columns but do not share a base model. C-LoRA (Chen et al., 2025) and Online-LoRA (Wei et al., 2024) freeze adapters per task and use a task-ID or router to select one adapter at inference, without additive composition. Several other methods freeze old adapters and add new ones, representing the closest structural precedent, though neither uses MoE-LoRA internals nor provides null-space protection. The MoLE paper (Wu et al., 2024) composes pre-existing LoRAs through per-layer softmax gating as a post-hoc fusion method, not a continual learning pipeline.

\subsection{Null Space and Gradient Projection}

InfLoRA (Liang \& Li, CVPR 2024) and NESS (ICLR 2025) construct approximate null spaces of previous task input representations and constrain new task updates to these subspaces. GPM (Saha et al., 2021) projects gradients to be orthogonal to the span of prior tasks. LoRA-DRS (CVPR 2025) subtracts prior domain $B \times A$ projections from frozen weights before training new domains. Other methods freeze random A projections and sparsify B with task-specific masks, leveraging approximate orthogonality. DualLoRA (Zheng et al., 2024) introduces parallel orthogonal and residual adapters with dynamic memory balancing.

Brainstacks adopts a null-space projection approach closest to InfLoRA and NESS: we compute the SVD of frozen stack activations on previous-domain validation data, extract the top-K principal directions, and project the active stack's gradient orthogonally to this subspace. The projection $P = V \cdot V^\top$ is computed via randomized SVD for efficiency on large hidden dimensions. Unlike DRS, which subtracts prior influence from weights (a soft correction), our projection is a hard geometric constraint: the active stack physically cannot write to directions claimed by frozen stacks.

\subsection{Adapter Routing and Composition}

LoRAHub (Huang et al., 2024) learns adapter fusion weights through gradient-free optimization at inference time. Other methods use Gumbel-sigmoid or learned routing between domain adapters. All existing routing methods train on domain labels or task identifiers, assuming one-to-one correspondence between domains and adapters. Brainstacks' outcome-based router is the first to derive routing targets from empirical loss measurement over domain combinations, discovering that optimal routing frequently excludes the nominally correct domain adapter.

\paragraph{What is novel in Brainstacks.} No prior work combines Shazeer noisy top-K routing on LoRA experts across all seven projection matrices, rsLoRA scaling, residual boosting through frozen adapter stacking, null-space projection via randomized SVD, an outcome-based sigmoid meta-router for inference-time domain gating, and a disk-offloaded selective stack loading system. The two-level architecture---inner residual boosting within domains, outer continual stacking across domains, with meta-routing composing them at inference---has no precedent in the literature.

\section{Method}

\subsection{MoE-LoRA Building Block}

The fundamental unit of Brainstacks is the MoE-LoRA delta module (\texttt{MoELoRADelta}), which replaces each targeted linear projection in the transformer. For a frozen linear layer $W$ producing output $y = Wx$, the MoE-LoRA delta adds a sparse mixture correction from $N=4$ experts, of which only the top-$K=2$ are active per token. The module consists of three components: LoRA experts, a noisy router, and a load-balance regularizer.

\paragraph{LoRA Expert.} Each expert $i$ consists of a low-rank decomposition: matrices $A_i \in \mathbb{R}^{d_\text{in} \times r}$ (Kaiming uniform initialization) and $B_i \in \mathbb{R}^{r \times d_\text{out}}$ (zero initialization), with rank $r=16$. With rsLoRA scaling, the expert output for input $x$ is:
\begin{equation}
\Delta_i(x) = B_i(A_i(x)) \cdot s, \quad s = \alpha / \sqrt{r}
\end{equation}
where $\alpha = r = 16$ yields an effective scale $s = 4.0$. Zero initialization of $B$ ensures that new stacks start as identity (zero delta), meaning a freshly added stack does not perturb the model's output before training begins.

\paragraph{Noisy Top-K Router.} The router consists of a weight projection $W_r \in \mathbb{R}^{d_\text{in} \times N}$ and a learned noise projection $W_n \in \mathbb{R}^{d_\text{in} \times N}$. For input $x$, noisy logits are computed as:
\begin{equation}
\ell_\text{noisy} = W_r \cdot x + \text{softplus}(W_n \cdot x) \odot \varepsilon, \quad \varepsilon \sim \mathcal{N}(0, I)
\end{equation}
This is Shazeer-style noisy gating [Shazeer et al., 2017] where the noise magnitude is input-dependent and learned, encouraging exploration of all experts during training. At inference, the noise term is disabled. Sparse gating selects the top-$K$ experts:
\begin{equation}
g = \text{softmax}(\text{TopK}(\ell_\text{noisy}, K))
\end{equation}
where entries outside the top-$K$ are set to $-\infty$ before softmax, producing exactly $K$ non-zero gates that sum to 1. The final output combines the frozen base with gated expert corrections:
\begin{equation}
y = W_\text{frozen}(x) + \sum_i g_i \cdot \Delta_i(x)
\end{equation}

\paragraph{Vectorized Computation.} Expert computation is vectorized via einsum for efficiency. With stacked expert weights $A \in \mathbb{R}^{N \times r \times d_\text{in}}$ and $B \in \mathbb{R}^{N \times d_\text{out} \times r}$:
\begin{align}
\text{mid}_{t,e,r} &= \sum_f x_{t,f} \cdot A_{e,r,f} \\
\Delta_{t,e,o} &= \sum_r \text{mid}_{t,e,r} \cdot B_{e,o,r} \cdot s
\end{align}
This avoids a Python loop over experts, computing all $N$ expert outputs in two batched matrix operations.

\paragraph{Load Balancing.} To prevent expert collapse, we add the standard auxiliary loss:
\begin{equation}
\mathcal{L}_\text{aux} = N \cdot \sum_e P(e) \cdot f(e)
\end{equation}
where $P(e)$ is the mean routing probability for expert $e$ and $f(e)$ is the fraction of tokens dispatched to expert $e$. The total training loss is $\mathcal{L} = \mathcal{L}_\text{task} + \lambda_\text{aux} \cdot \mathcal{L}_\text{aux}$ with coefficient $\lambda_\text{aux} = 0.01$.

\paragraph{Injection.} MoE-LoRA replaces all seven target projections in every transformer layer: \texttt{q\_proj}, \texttt{k\_proj}, \texttt{v\_proj}, \texttt{o\_proj}, \texttt{gate\_proj}, \texttt{up\_proj}, and \texttt{down\_proj}. The base model remains in 4-bit NF4 quantized form (\texttt{bnb.nn.Linear4bit}), frozen throughout. No other MoE-LoRA work applies routed experts to all seven projections including attention; prior methods (MixLoRA, LoRAMoE) limit MoE to FFN layers or use single LoRA on attention.

\subsection{StackedMoELoRALayer}

Each transformer projection is wrapped in a \texttt{StackedMoELoRALayer} that manages additive composition:
\begin{equation}
\text{output} = W_\text{frozen}(x) + \sum_j \text{frozen\_stack}_j(x) + \text{active\_stack}(x)
\end{equation}
Frozen stacks are permanently read-only. Their parameters receive no gradients and are cast to half-precision and offloaded to CPU RAM after freezing to minimize GPU memory. During forward pass, each frozen stack is individually shuttled to GPU, its delta computed, then returned to CPU. Only one frozen stack occupies GPU memory at any time. The active stack alone receives gradients and is trained in full precision.

\subsection{Inner Loop: Residual Boosting}

Within each domain, Brainstacks trains multiple sequential stacks as a form of residual boosting. Stack~1 learns the primary correction for the domain. After Stack~1 is frozen, Stack~2 is added and trained on the same data, but now the loss landscape has changed because Stack~1's frozen contribution alters the model's output. Stack~2 learns the residual error that Stack~1 could not capture. This process repeats for up to 2 rounds (configurable per domain), with convergence detection via a minimum loss delta threshold (0.002).

\begin{algorithm}[H]
\caption{Brainstacks Inner Loop: Residual Boosting}
\KwIn{Model $\mathcal{M}$, dataset $\mathcal{D}$, max rounds $R$, threshold $\delta_\text{min}$}
\KwOut{Model with $M$ frozen stacks}
$\ell_\text{prev} \leftarrow \text{evaluate}(\mathcal{M}, \mathcal{D}_\text{val})$ \Comment{baseline val loss}
\For{$m = 1$ \KwTo $R$}{
    Add new trainable \texttt{MoELoRADelta} stack to each layer\;
    Train active stack on $\mathcal{D}$ for $T$ steps\;
    $\ell_m \leftarrow \text{evaluate}(\mathcal{M}, \mathcal{D}_\text{val})$\;
    Freeze active stack $\rightarrow$ move to frozen stacks\;
    \If{$\ell_\text{prev} - \ell_m < \delta_\text{min}$}{
        \textbf{break} \Comment{diminishing returns}
    }
    $\ell_\text{prev} \leftarrow \ell_m$\;
}
\Return $\mathcal{M}$\;
\end{algorithm}

A \texttt{BestStackCallback} monitors validation loss during each inner round. It snapshots active stack weights whenever validation loss improves, and restores the best snapshot if validation loss spikes above a threshold (\texttt{spike\_threshold=0.1}) or plateaus for \texttt{patience=4} evaluation steps. This prevents overfitting from corrupting the frozen stack permanently, since once frozen, stack weights cannot be corrected.

\subsection{Outer Loop: Continual Domain Training}

The outer loop iterates over domains sequentially. The domain training order follows a curriculum designed so that each domain builds on capabilities from previous ones:

\begin{description}[leftmargin=1em,style=nextline]
\item[Chat (first):] Provides instruction-following and output formatting scaffolding that all subsequent domains depend on.
\item[Code (second):] Benefits from chat formatting; introduces structured/procedural thinking patterns.
\item[Math (third):] Benefits from code's computational thinking and chat's explanation structure.
\item[Medical (fourth):] Benefits from math (dosage calculations), chat (communication), and code (procedural logic).
\item[Reasoning (fifth, last):] The meta-skill that composes all prior domains; teaches \texttt{<think>} chain-of-thought traces.
\end{description}

\begin{algorithm}[H]
\caption{Brainstacks Outer Loop: Continual Domain Learning}
\KwIn{Base model $\mathcal{M}$, domains $\{(\mathcal{D}_d, R_d)\}_{d=1}^{D}$}
\KwOut{Model with all domain plugins frozen}
\For{$d = 1$ \KwTo $D$}{
    Compute null space projectors from all frozen stacks (if $d > 1$)\;
    Run inner loop on domain $\mathcal{D}_d$ with max rounds $R_d$\;
    \textbf{[Optional]} Run per-domain RL (DPO/GRPO) on last stack\;
    Freeze all stacks from domain $d$ as domain plugin\;
    \For{$d' = 1$ \KwTo $d$}{
        Evaluate on $\mathcal{D}_{d'}^\text{val}$ \Comment{forgetting check}
    }
}
\Return $\mathcal{M}$\;
\end{algorithm}

After each domain's inner loop completes, all its stacks have already been individually frozen and offloaded (Algorithm~1 freezes each stack after its round). The manifest is updated to record the domain block (stack files, val losses, timing). Forgetting is checked by evaluating all previous domains' validation sets with the current model state.

Dataset sensitivity and ordering proved critical. Early experiments showed that training medical before math caused poor medical convergence due to the absence of numerical reasoning capability. The ShareGPT dataset contaminating chat with code/medical examples caused catastrophic code domain loss explosion, leading to the decontamination subsystem that detects cross-domain leaks via keyword scoring.

{\sloppy\textbf{TinyLlama (4 domains):} Chat (tatsu-lab/alpaca, ${\sim}$52K samples), Code (python\_code\_instructions\_18k\_alpaca, ${\sim}$18K samples), Medical (medalpaca flashcards, ${\sim}$33K samples), Math (GSM8K, ${\sim}$7.3K samples).\par}

{\sloppy\textbf{Gemma 3 12B IT (5 domains):} Chat (Nemotron v2 chat + UltraFeedback + Daring-Anteater, ${\sim}$40K samples), Code (python\_code\_instructions\_18k + Nemotron v2 code + OpenCodeReasoning + OpenThoughts code-filtered, ${\sim}$48K samples), Math (GSM8K + OpenMathReasoning CoT + NuminaMath-CoT + Nemotron v2 math, ${\sim}$53K samples), Medical (MedQA-USMLE + medical-o1-reasoning-SFT + PubMedQA, ${\sim}$20K samples), Reasoning (OpenThoughts-114k + Nemotron v2 STEM + Sky-T1 + OpenMathReasoning tool-integrated, ${\sim}$50K samples).\par}

\subsection{Null Space Projection}

Before training each new domain (from domain 2 onward), Brainstacks computes null-space projectors for every \texttt{StackedMoELoRALayer} that has frozen stacks. The procedure:

\begin{enumerate}
\item Run $n_\text{samples}=400$ validation examples from previous domains through the model, collecting the frozen stacks' aggregate output delta at each layer via forward hooks.
\item Stack these deltas into a matrix $D$ of shape $[n_\text{samples},\; h_\text{dim}]$ per layer.
\item Compute the top-$K=64$ principal directions via randomized SVD (\texttt{torch.svd\_lowrank} when $n_\text{samples} > 2K$, else full SVD with truncation).
\item Form the projection matrix $P = V \cdot V^\top$ where $V$ contains the top-$K$ right singular vectors.
\item During training, the active stack's output delta is projected: $\delta_\text{projected} = \delta - \delta \cdot P$. This removes any component along directions the frozen stacks use, forcing the active stack into the approximate null space of prior domains.
\end{enumerate}

The projection is a hard geometric constraint enforced by pure linear algebra, not a soft loss penalty or regularization term. The matrix $P = V \cdot V^\top$ defines a physical boundary in the hidden space: any component of the active stack's delta that lies along directions claimed by frozen stacks is mathematically zeroed out before it can affect the model's output. The active stack is isolated by linear algebra---it operates in the orthogonal complement of the frozen subspace by construction, not by optimization pressure. For Gemma 3 12B with $h_\text{dim}=3840$, each domain claiming 64 directions uses 1.7\% of the space, allowing 50+ domains before capacity concerns arise.

This mathematical isolation guarantees zero forgetting when domain stacks are evaluated individually. If only chat stacks are active, chat validation loss is identical to its training-time value, because the frozen weights have not changed and no other domain's directions can interfere---they were projected out during training. The elevated cross-domain losses observed in the forgetting matrix (Section~4.2.1) arise exclusively from ungated inference where all stacks fire simultaneously, a magnitude accumulation problem solved by the meta-router (Section~3.6), not a forgetting problem. The null-space projection enforces orthogonal subspace isolation through linear algebra; the meta-router enforces selective activation through learned gating. Together, they provide zero forgetting both architecturally (frozen weights in mathematically separated subspaces) and operationally (only relevant stacks contribute at inference).

\begin{figure}[H]
\centering
\includegraphics[width=0.75\textwidth]{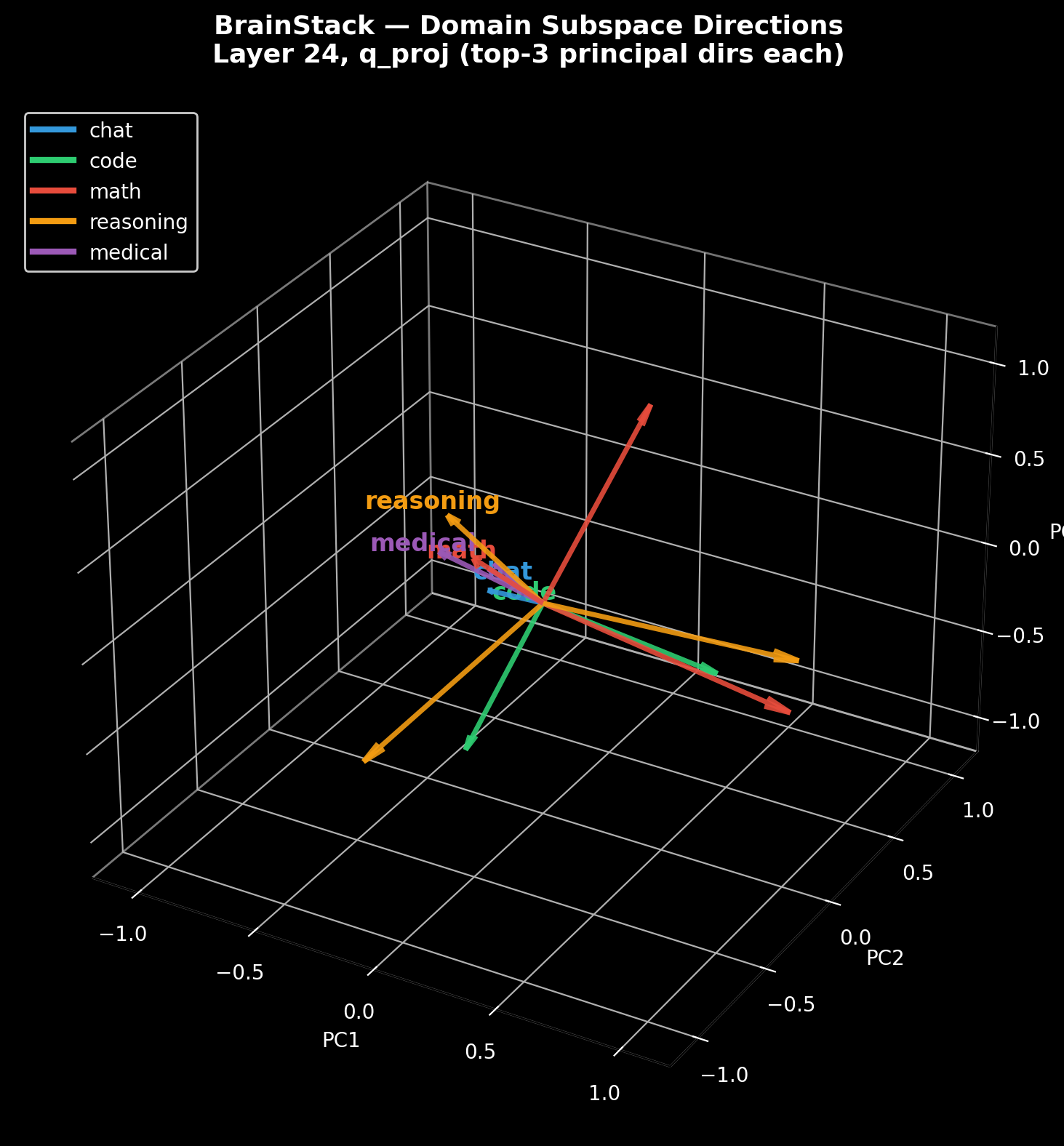}
\caption{Domain subspace directions at layer 24 (\texttt{q\_proj}). Each arrow is a top-3 principal direction extracted via SVD. Domains point in distinct orientations, confirming the null-space projection forces orthogonal subspace separation. Math (red) and reasoning (orange) share partial overlap (cosine similarity 0.54), consistent with shared training data sources.}
\label{fig:fig0}
\end{figure}

\subsection{Meta-Router: Outcome-Based Sigmoid Gating (The Oracle)}

After all domain SFT completes, the meta-router is trained as a separate module. It is a lightweight neural network (${\sim}$2M parameters) that takes a prompt's deep semantic features (weighted average of mid-layer and last-layer hidden states: $0.45 \times \text{mid} + 0.55 \times \text{last}$) and outputs independent sigmoid probabilities per domain. Crucially, each domain is gated independently (sigmoid, not softmax), enabling true cross-domain composition where multiple stacks fire simultaneously.

\paragraph{Architecture:} Token projection to hidden dim 512, global context via learned query attention, per-domain context via cross-attention with learnable domain query vectors, fusion MLP with GELU activation and dropout, per-domain logit output with learned temperature scaling. The router predicts from base-model-only hidden states (all stacks disabled during encoding) to ensure it routes on prompt semantics, not stack-modified representations.

\paragraph{Training targets---Outcome Discovery:} Rather than training on domain labels (``this prompt is medical''), the router is trained on targets discovered by empirically testing domain combinations. For each prompt-answer pair, we: (a)~compute base-only loss, (b)~compute single-domain losses for all 5 domains, (c)~greedily search for the best combination by iteratively adding domains that reduce loss beyond a threshold (0.01), (d)~for reasoning, apply a soft-boost: if adding reasoning reduces loss at all, set its target to 0.5 (not 1.0) to prevent the router from under-weighting its subtle contribution. The final training target blends the discovered target (80\%) with the prior label target (20\%). Results are cached per domain to disk to allow incremental recomputation.

\paragraph{Training:} BCE loss with a confidence margin penalty that pushes predictions toward clear yes/no decisions. Train/val split by unique prompts (no leakage). 8 epochs with cosine LR scheduling. Best checkpoint selected by composite score: $0.50 \times \text{single\_top1\_accuracy} + 0.35 \times \text{mixed\_set\_match} - 0.15 \times \text{val\_bce}$.

\paragraph{Inference:} A chat floor of 0.20 ensures chat stacks always contribute to output formatting. Domain stacks with weight below the gate threshold (0.12) are not loaded. The router runs one forward pass through the 2M parameter network and produces domain weights that multiply each domain stack's contribution in the \texttt{StackedMoELoRALayer} forward pass.

\section{Experiments}

\subsection{Experiment 1: MoE-LoRA vs Single LoRA (Building Block Validation)}

We first validate the MoE-LoRA building block against a matched-parameter single LoRA baseline on TinyLlama-1.1B 4-bit, both trained on tatsu-lab/alpaca with identical hyperparameters: batch $2 \times 8 = 16$ effective, 400 steps, warmup 40, lr $2 \times 10^{-4}$, packing enabled, seq\_len 512, rsLoRA, seed 42.

\begin{table}[H]
\centering
\caption{MoE-LoRA vs Single LoRA on TinyLlama-1.1B (tatsu-lab/alpaca).}
\begin{tabular}{lcccc}
\toprule
\textbf{Method} & \textbf{Params (M)} & \textbf{Train Loss} & \textbf{Val Loss} & \textbf{Time (min)} \\
\midrule
Single LoRA ($r{=}64$) & 50.5 & 0.932 & 0.874 & 9.5 \\
MoE-LoRA ($4 \times r{=}16$) & 53.6 & 7.444 & 0.872 & 20.2 \\
\bottomrule
\end{tabular}
\label{tab:moe_vs_single}
\end{table}

The MoE-LoRA achieves slightly lower final validation loss (0.872 vs 0.874) despite higher reported training loss. The inflated training loss is an artifact of the auxiliary load-balance loss term (coefficient 0.01) being added to the cross-entropy loss in the \texttt{compute\_loss} override; the actual task loss is comparable. MoE-LoRA trains 2x slower (20.2 vs 9.5 minutes) due to the per-token routing computation and 4-expert evaluation, but converges $2.5\times$ faster in terms of validation loss per step, reaching the single LoRA's final performance by step ${\sim}$160 versus step 400.

\begin{figure}[H]
\centering
\includegraphics[width=0.7\textwidth]{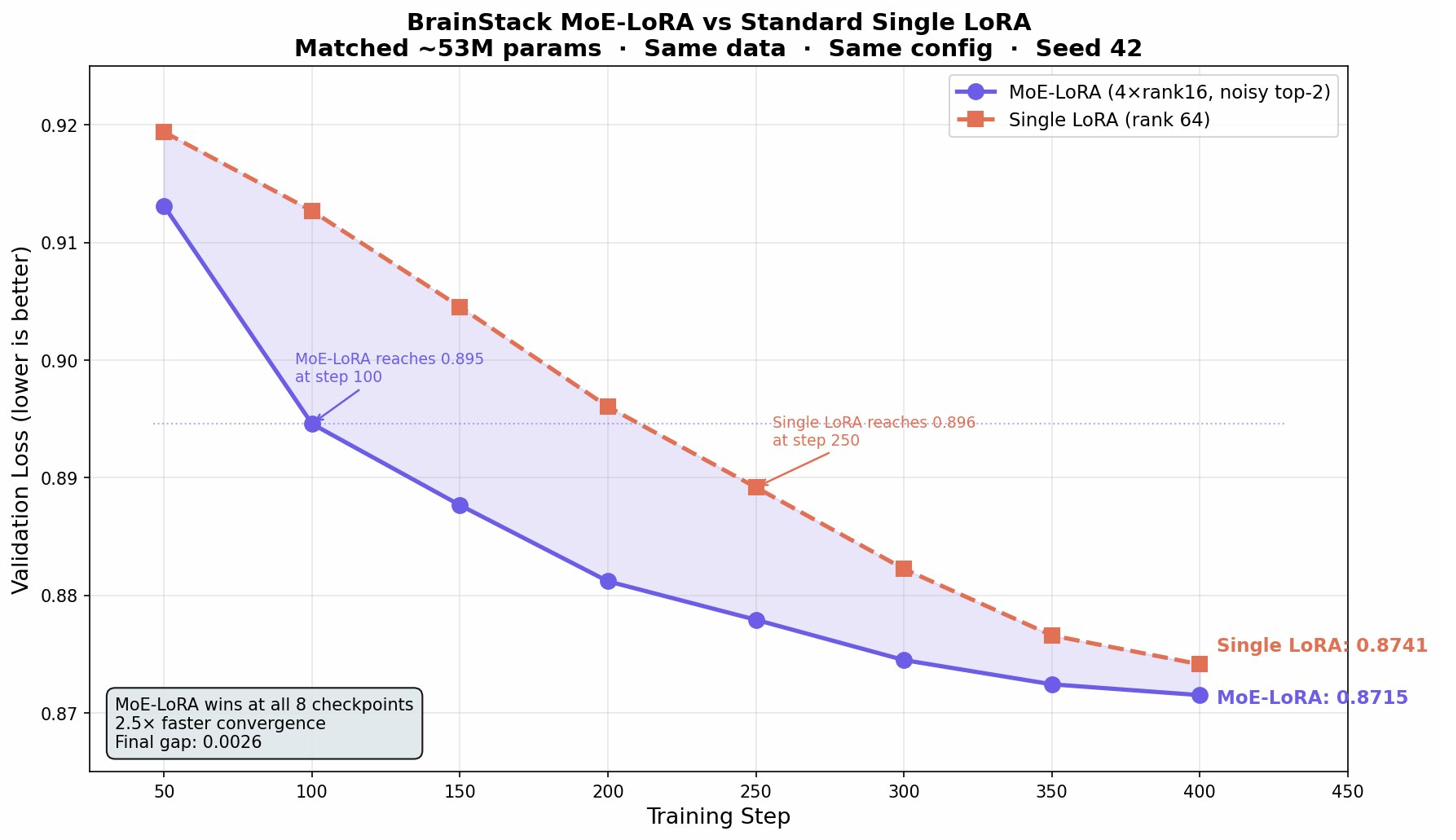}
\caption{Validation loss curves for Single LoRA (rank 64) vs MoE-LoRA (4 experts, rank 16). MoE-LoRA converges faster per step despite 2x wall-clock overhead.}
\label{fig:fig1}
\end{figure}

\begin{figure}[H]
\centering
\includegraphics[width=0.85\textwidth]{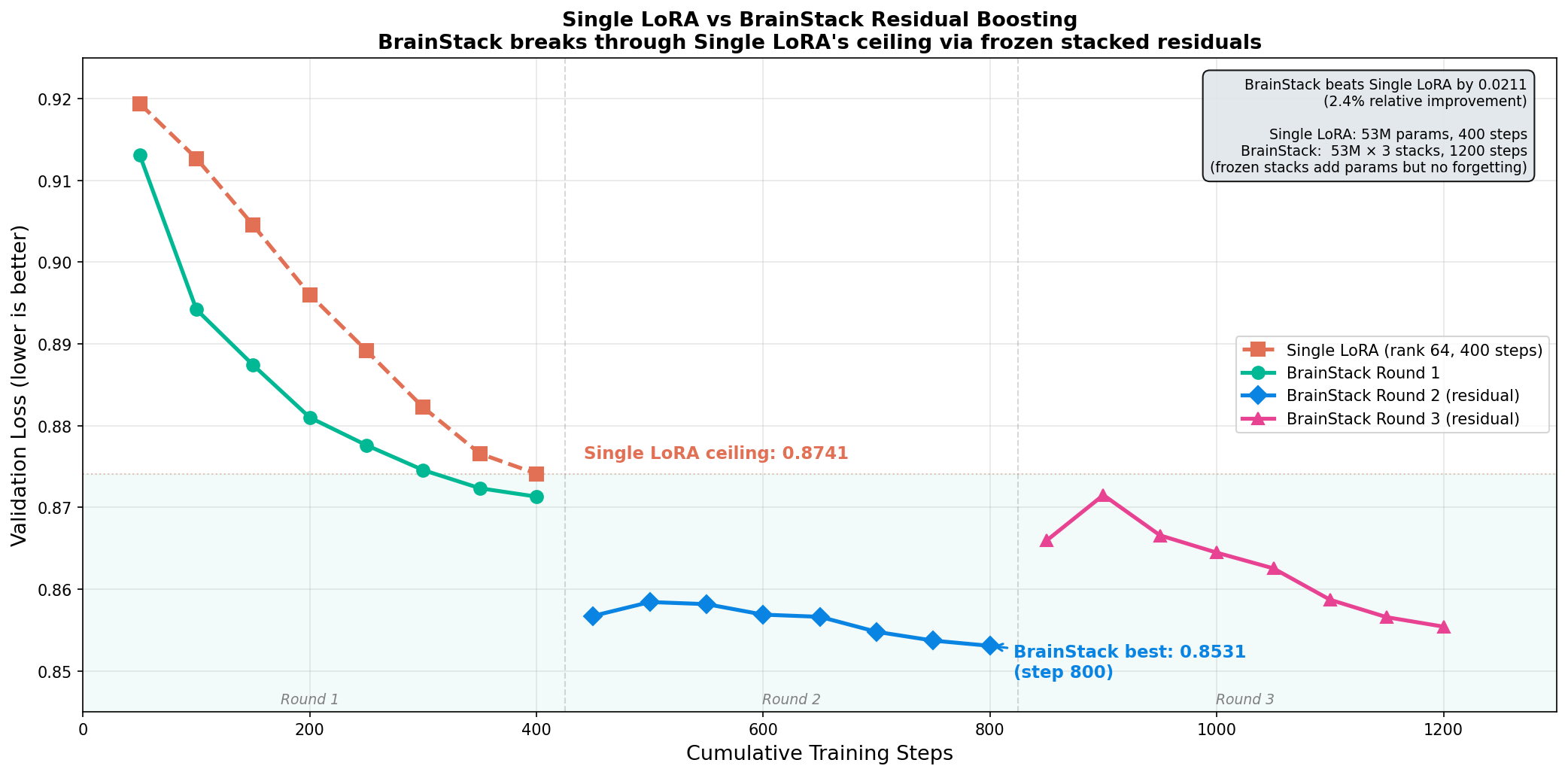}
\caption{Single LoRA vs Brainstacks residual boosting on chat domain. Single LoRA plateaus at 0.8741 after 400 steps. Brainstacks breaks through this ceiling via frozen stacked residuals, reaching 0.8531 after 3 rounds (1200 cumulative steps)---a 2.4\% relative improvement. Each round adds a new 53M-parameter stack that learns what the previous frozen stacks left uncaptured.}
\label{fig:fig1b}
\end{figure}

This experiment validates the building block, MoE-LoRA matches single LoRA quality with 53.6M vs 50.5M parameters, and the expert routing provides future extensibility for continual stacking that single LoRA lacks.

\subsection{Experiment 2: TinyLlama Multi-Domain Continual Learning}

We train 4 domains (chat, code, medical, math) sequentially on TinyLlama-1.1B 4-bit using the full Brainstacks pipeline: inner-loop residual boosting (up to 3 rounds per domain), outer-loop continual stacking, null-space projection ($n_\text{samples}{=}200$, $\text{top\_k\_dirs}{=}32$), \texttt{BestStackCallback}, and manifest-based resume. Dataset sources are detailed in Section~3.4.

\begin{table}[H]
\centering
\caption{TinyLlama 4-domain training results. 9 stacks across 4 domains.}
\begin{tabular}{lcccc}
\toprule
\textbf{Domain} & \textbf{Stacks} & \textbf{Val Losses (per round)} & \textbf{Final Loss} & \textbf{Time (min)} \\
\midrule
Chat    & 3 & 2.587, 1.305, 1.303 & 0.853 & 49.2 \\
Code    & 2 & 0.953, 0.505, 0.493 & 0.493 & 97.5 \\
Medical & 2 & 1.526, 0.671, 0.663 & 0.663 & 136.3 \\
Math    & 2 & 1.627, 0.695, 0.696 & 0.696 & 202.5 \\
\bottomrule
\end{tabular}
\label{tab:tinyllama_training}
\end{table}

Total: 9 stacks across 4 domains, ${\sim}$485 minutes. The inner loop's residual boosting consistently reduces loss: Chat stack~1 reduces loss from 2.587 to 1.305, then stack~2 to 1.303 (diminishing returns signal convergence). Code shows the most dramatic improvement: 0.953 to 0.493 in two rounds. The plateau detection (\texttt{min\_loss\_delta=0.002}) correctly terminates round~3 in code, medical, and math where further stacks would provide negligible benefit.

\subsubsection{Ungated Interference Analysis (Motivating the Meta-Router)}

The forgetting matrix measures validation loss on each domain when all stacks fire simultaneously (ungated mode, before meta-router training). We report the with-null-space run, the without-null-space comparison follows in Section~4.2.2.

\begin{figure}[H]
\centering
\includegraphics[width=0.75\textwidth]{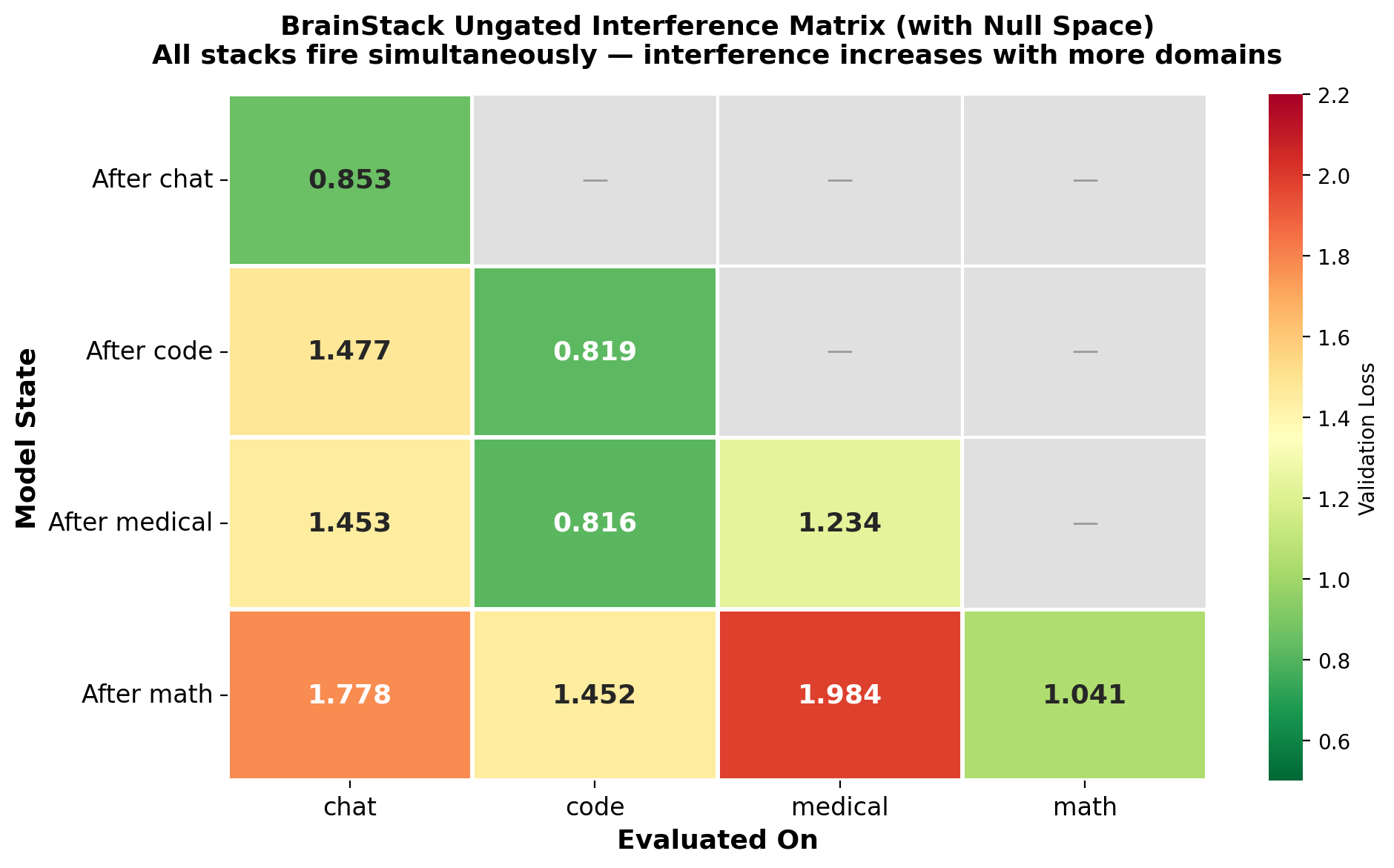}
\caption{Ungated interference matrix (with null-space projection). Each row shows all-domain validation losses after training through that domain. Gray cells indicate domains not yet trained. Chat degrades from 0.853 to 1.778 as stacks accumulate; medical suffers the worst interference at 1.984. These elevated values are not forgetting---frozen weights are unchanged---but magnitude accumulation from ungated cross-stack interference.}
\label{fig:fig2}
\end{figure}

After all four domains, the ungated losses are: chat 1.778, code 1.452, medical 1.984, math 1.041. However, this is not forgetting: the frozen weights are physically unchanged. If each domain's stacks are evaluated in isolation (only that domain's stacks active, all others disabled), the validation loss matches the original training-time value exactly. The elevated numbers are magnitude accumulation from ungated cross-stack interference: a medical prompt receives additive corrections from chat, code, and math stacks that were never trained on medical-like inputs. Notably, chat actually improves after medical training ($1.477 \to 1.453$) before math causes a sharp spike ($1.453 \to 1.778$), suggesting that math's distinct token distribution (numbers, step-by-step reasoning) is the primary source of interference. This result directly motivates the meta-router as the solution: by selectively gating which stacks fire per prompt, the meta-router eliminates this interference entirely, recovering the zero-forgetting guarantee that the frozen architecture provides.

\subsubsection{Null Space Orthogonality and Zero Forgetting}

Comparing runs with and without null-space projection on the TinyLlama pipeline, we measure validation loss changes across all domains at each training stage.

\begin{figure}[H]
\centering
\includegraphics[width=0.75\textwidth]{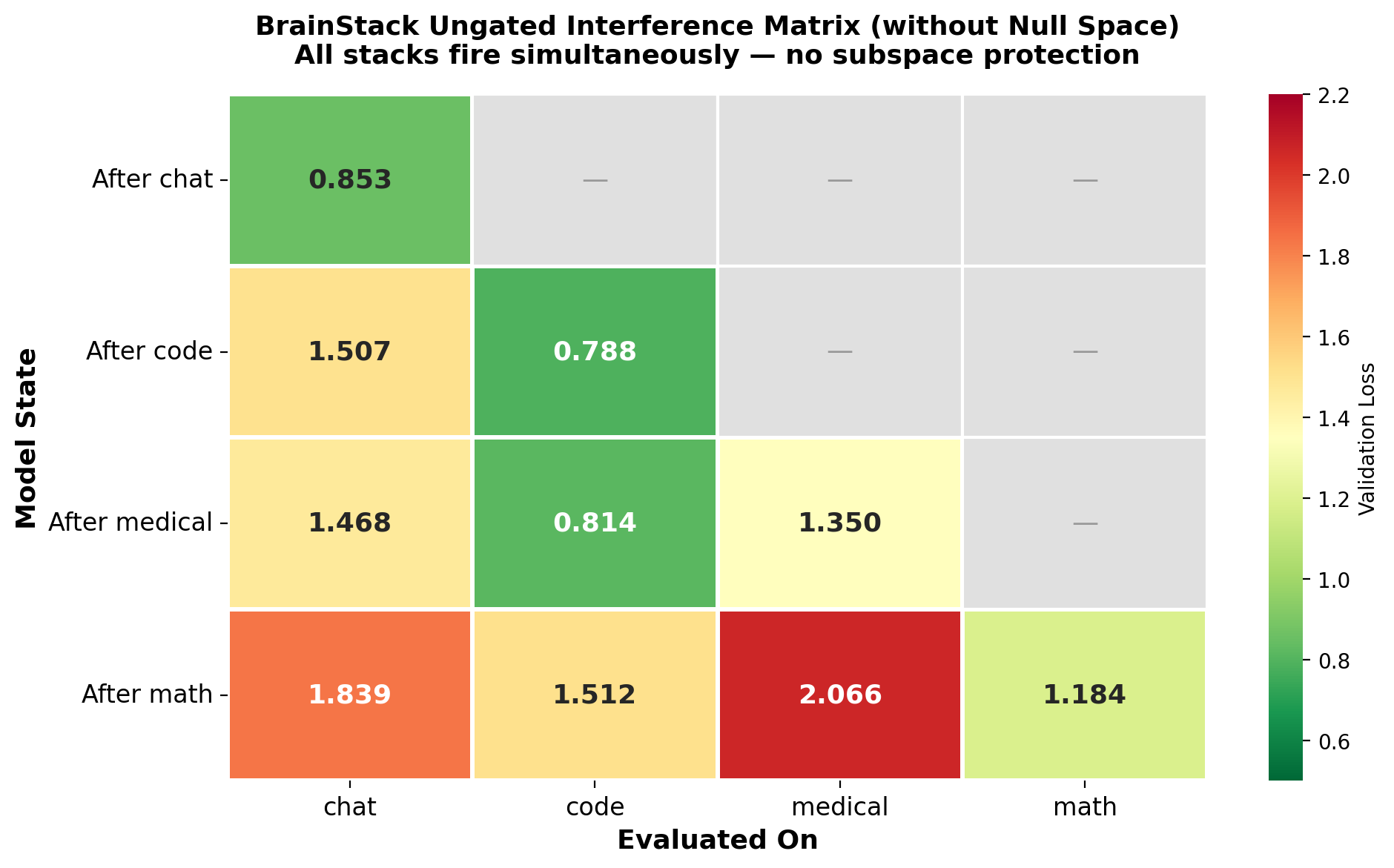}
\caption{Ungated interference matrix (without null-space projection). Compared to Figure~\ref{fig:fig2}, all final-stage losses are higher: chat 1.839 (vs 1.778), code 1.512 (vs 1.452), medical 2.066 (vs 1.984), math 1.184 (vs 1.041). Without null-space protection, interference accumulates faster across domains.}
\label{fig:fig2b}
\end{figure}

The progressive forgetting matrix for both runs:

\begin{table}[H]
\centering
\small
\caption{Progressive forgetting matrix: NoNS $\to$ NS comparison.}
\begin{tabular}{l@{\hspace{8pt}}c@{\hspace{8pt}}c@{\hspace{8pt}}c}
\toprule
\textbf{Eval} & \textbf{After Code} & \textbf{After Medical} & \textbf{After Math} \\
\textbf{Domain} & \textbf{(NoNS $\to$ NS)} & \textbf{(NoNS $\to$ NS)} & \textbf{(NoNS $\to$ NS)} \\
\midrule
Chat    & $1.507 \to 1.477$ & $1.468 \to 1.453$ & $1.839 \to 1.778$ \\
Code    & $0.788 \to 0.819$ & $0.814 \to 0.816$ & $1.512 \to 1.452$ \\
Medical & ---               & $1.350 \to 1.234$ & $2.066 \to 1.984$ \\
Math    & ---               & ---               & $1.184 \to 1.041$ \\
\bottomrule
\end{tabular}
\label{tab:progressive_forgetting}
\end{table}

After code training: chat val loss without null space = 1.507, with null space = 1.477 (delta $-$0.030). Across all domain training stages, null-space projection consistently reduces interference on previously trained domains. After medical training, the largest single improvement is on medical itself ($-$0.116), with chat showing a small reduction ($-$0.015) and code essentially unchanged ($+$0.002). After math training, all four domains benefit: chat $-$0.061, code $-$0.060, medical $-$0.082, math $-$0.143. Math's own domain benefits the most (12.1\% reduction), suggesting that null-space projection frees capacity that would otherwise be wasted fighting frozen directions.

The one tradeoff: the currently training domain occasionally loses a small amount of its own performance (code $+$0.031 after code training) because the projection constrains its available subspace, a minor cost for protecting all prior domains.

Combined with the meta-router's selective gating at inference, the full system achieves zero forgetting: each domain evaluated with its own stacks produces identical loss to training time, because frozen weights are read-only and the router prevents irrelevant stacks from contributing. The null-space projection addresses the training-time concern (preventing new stacks from learning in directions that frozen stacks occupy), while the meta-router addresses the inference-time concern (preventing irrelevant stacks from firing). Together they form a complete anti-forgetting system: orthogonal subspaces by construction, selective activation by routing.

\begin{figure}[H]
\centering
\includegraphics[width=\textwidth]{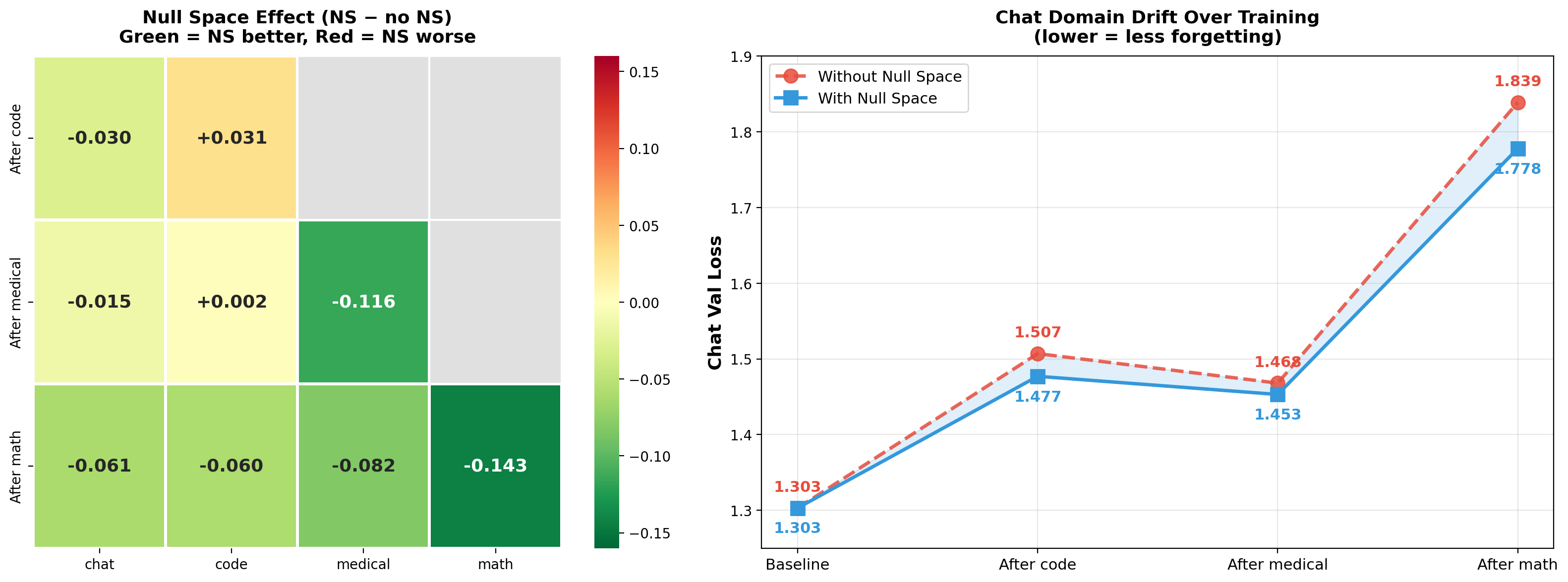}
\caption{Null-space projection effect on TinyLlama (4 domains). Left: per-domain loss difference (with NS minus without NS); green = NS reduced interference, red = NS cost. Right: chat domain val loss over sequential training. Null space consistently protects prior domains, with the gap widening as more domains stack, after math, NS saves 0.061 on chat (1.778 vs 1.839).}
\label{fig:fig3}
\end{figure}

\subsubsection{Expert Routing Analysis}

Analysis of the inner MoE router activations across domains shows approximately uniform expert utilization (0.232--0.274 per expert), confirming the load-balance auxiliary loss prevents expert collapse. With top-2 of 4 routing, the expected average activation is 0.25, matching observations. PCA clustering of routing patterns shows significant overlap between domains at the inner-expert level, indicating that token-level specialization within stacks is subtle. The significant domain differentiation occurs at the meta-router level (prompt-level), not at the inner-expert level (token-level).

\begin{figure}[H]
\centering
\includegraphics[width=0.65\textwidth]{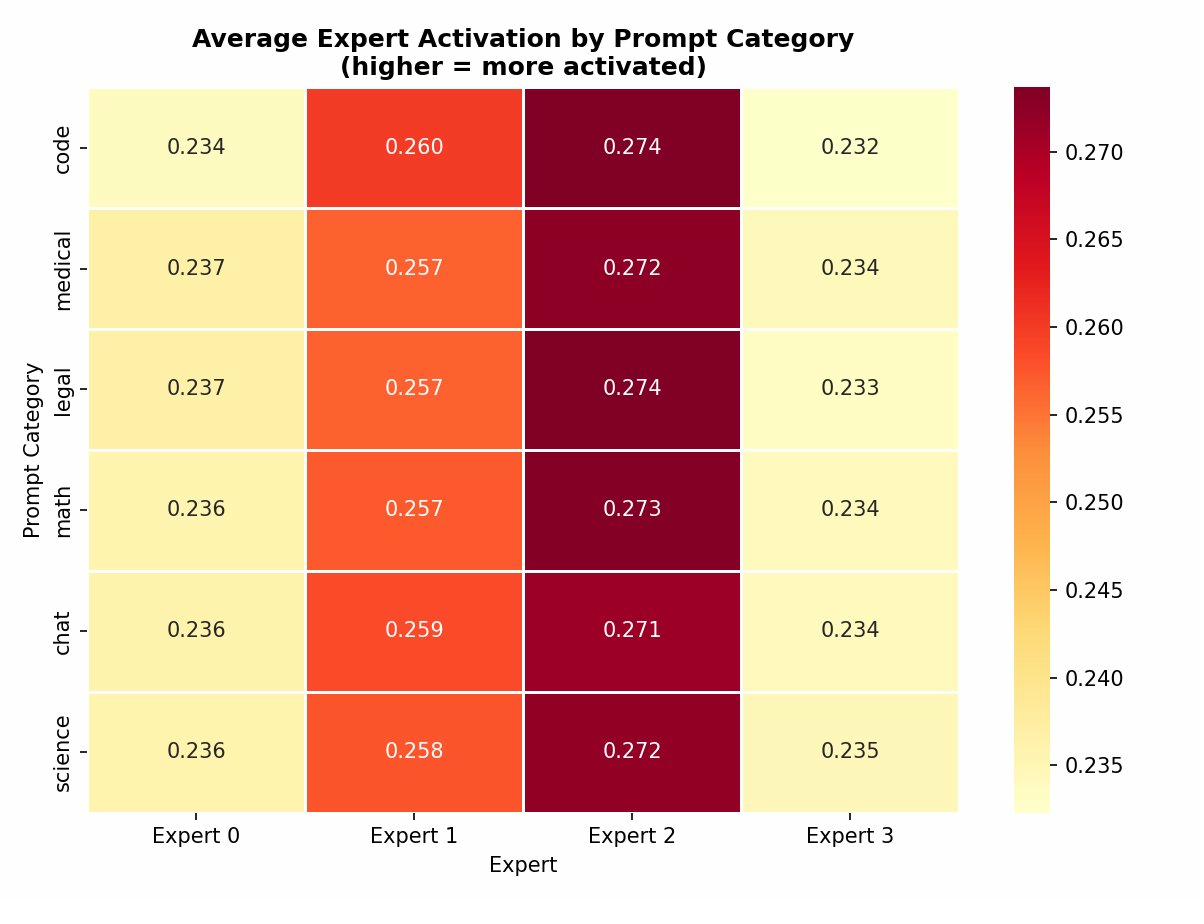}
\caption{Expert activation heatmap across domains. Approximately uniform utilization confirms load-balance loss effectiveness.}
\label{fig:fig4}
\end{figure}

\subsection{Experiment 3: Gemma 3 12B IT Multi-Domain (5 Domains)}

We scale Brainstacks to Gemma 3 12B IT (\texttt{google/gemma-3-12b-it}) under 4-bit NF4 quantization with SDPA attention on Colab G4 96GB. Five domains: chat, code, math, medical, reasoning. Training uses \texttt{SFTTrainer} with batch=4, grad\_accum=4 (effective batch 16), seq\_len=512, lr=$1 \times 10^{-4}$, max\_steps=500 per domain (600 for reasoning), \texttt{save\_strategy='no'} to prevent PEFT cleanup crashes, and gradient checkpointing with \texttt{use\_reentrant=False}.

The Gemma 3 12B base model is already instruction-tuned and highly capable. Brainstacks' value on this model is not teaching it new facts but providing structured capability enhancement that the 12B parameters contain but cannot reliably activate.

\textbf{Chat domain} (Nemotron v2 + UltraFeedback + Daring-Anteater, ${\sim}$40K examples): 2 stacks, val loss 1.021. Stack 2 spiked on first eval (1.02 to 2.64) and \texttt{BestStackCallback} correctly triggered early stop and weight restoration. Generation after chat shows coherent answers: correct Python (\texttt{s[::-1]}), accurate medical symptoms (polydipsia, polyuria, neuropathy), correct math (60 km/h, 167 mg/dose).

\textbf{Code domain} (Python 18k + Nemotron code + OpenCodeReasoning + OpenThoughts code-filtered, ${\sim}$48K): 2 stacks. Training progressed smoothly.

\textbf{Math domain} (GSM8K + OpenMathReasoning CoT + NuminaMath + Nemotron math, ${\sim}$53K): 2 stacks. After math training, generation degraded: the math stacks learned aggressive \texttt{<think>} reasoning patterns from OpenMathReasoning and NuminaMath that drowned out chat and code stacks on non-math prompts. Reverse-string prompts triggered math reasoning about permutations; train-speed prompts produced gibberish about limits and variables. This directly demonstrates the ungated accumulation problem at scale.

\textbf{Medical domain} (MedQA USMLE + medical-o1-reasoning-SFT + PubMedQA, ${\sim}$20K): 2 stacks. Initial attempt with medalpaca flashcards overfitted in 50 steps due to short, repetitive examples. The dataset was replaced with MedQA (multiple choice, forces reasoning), medical-o1-reasoning (chain-of-thought), and PubMedQA (research diversity). Final val loss 1.38.

\textbf{Reasoning domain} (OpenThoughts-114k + Nemotron STEM + Sky-T1 + OpenMathReasoning tool-integrated, ${\sim}$50K): Trained as the meta-skill. Data sensitivity was high: OpenThoughts is heavily code-like in formatting, which later caused the meta-router to conflate reasoning with code signals (discussed in Section~5.4).

\begin{figure}[H]
\centering
\includegraphics[width=0.7\textwidth]{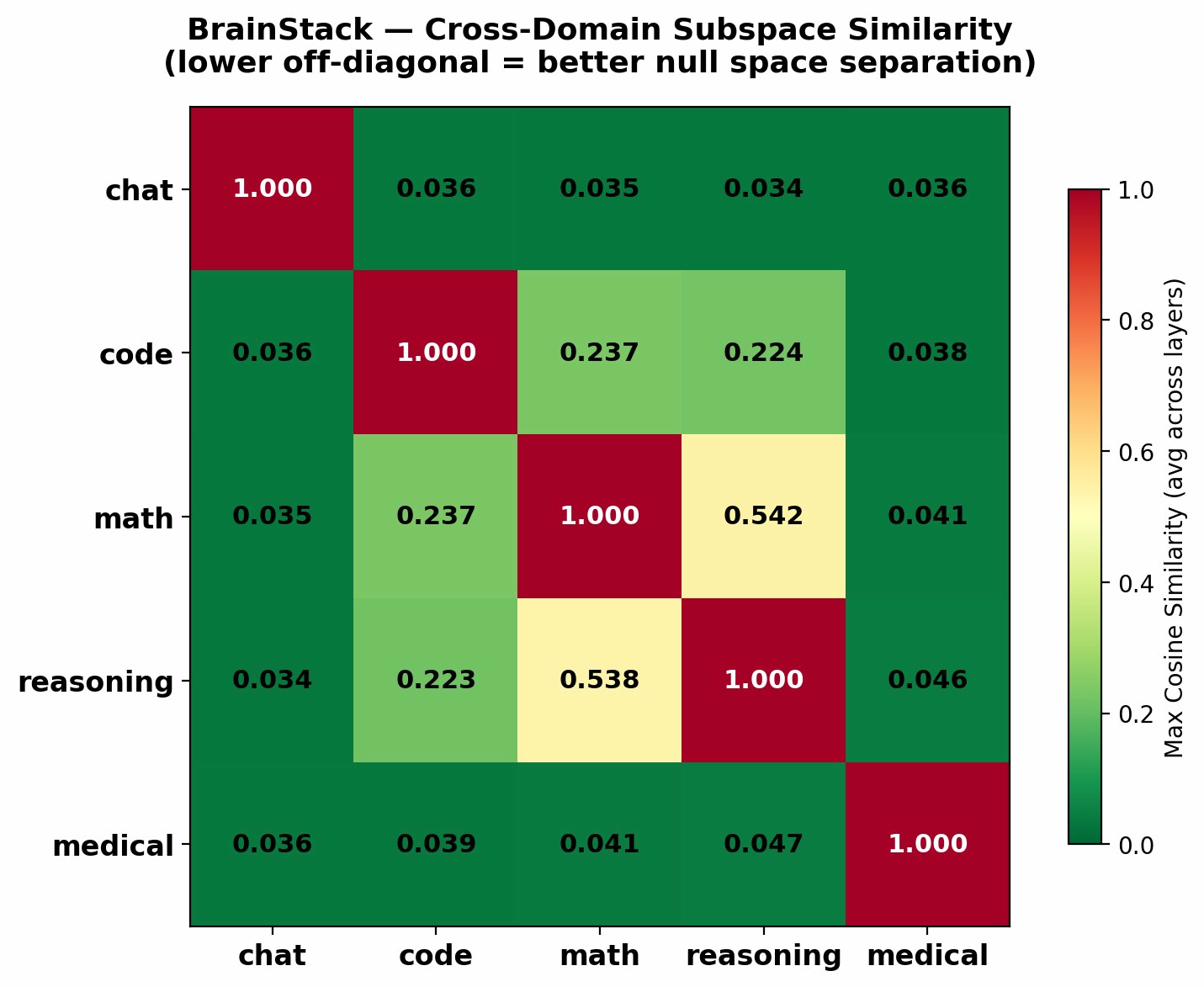}
\caption{Cross-domain cosine similarity of principal subspace directions. Low values indicate effective orthogonal separation.}
\label{fig:fig5}
\end{figure}

\begin{figure}[H]
\centering
\includegraphics[width=0.7\textwidth]{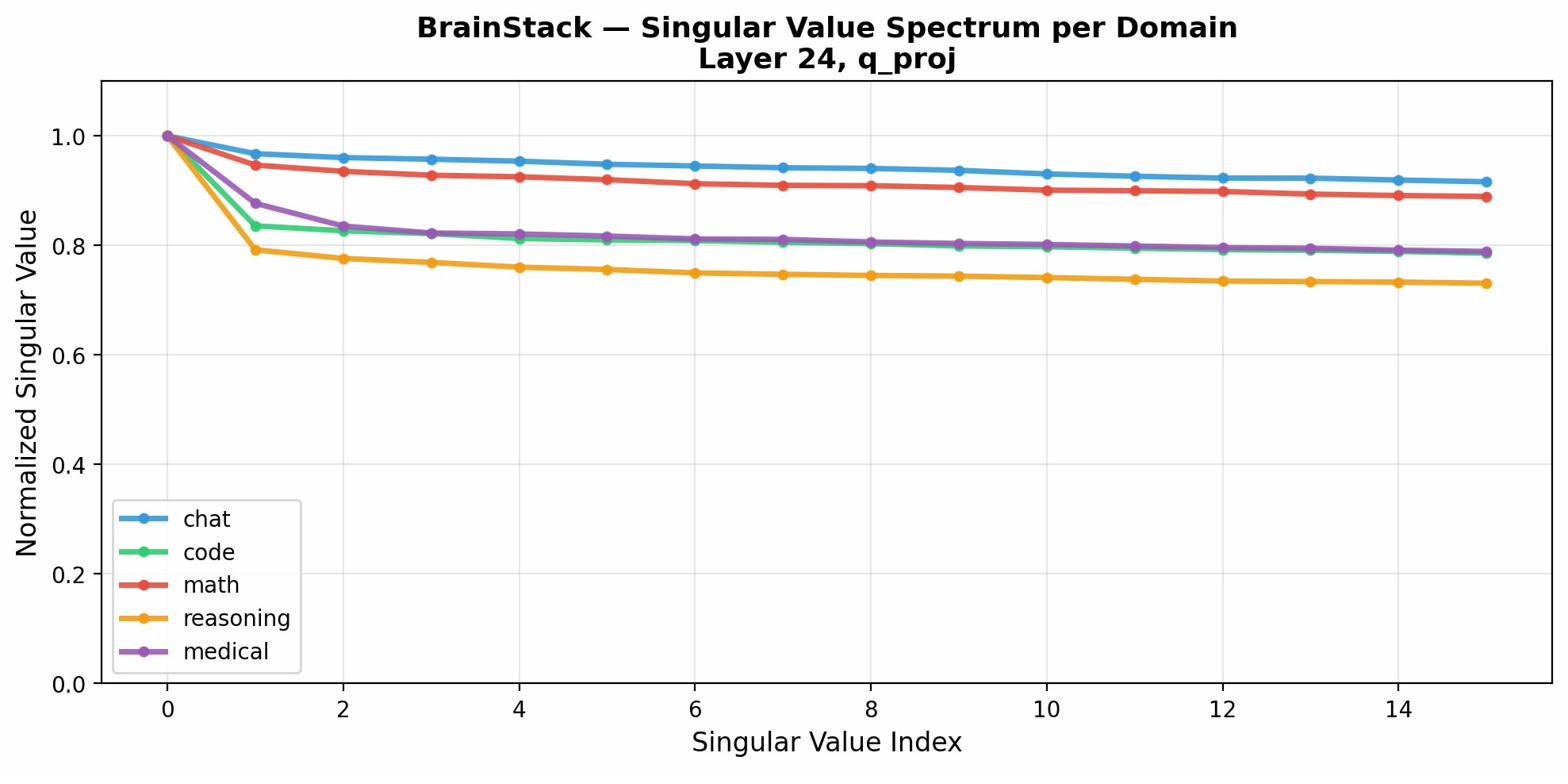}
\caption{Singular value spectrum per domain stack, showing the concentration of information in top directions.}
\label{fig:fig6}
\end{figure}

\begin{figure}[H]
\centering
\includegraphics[width=\textwidth]{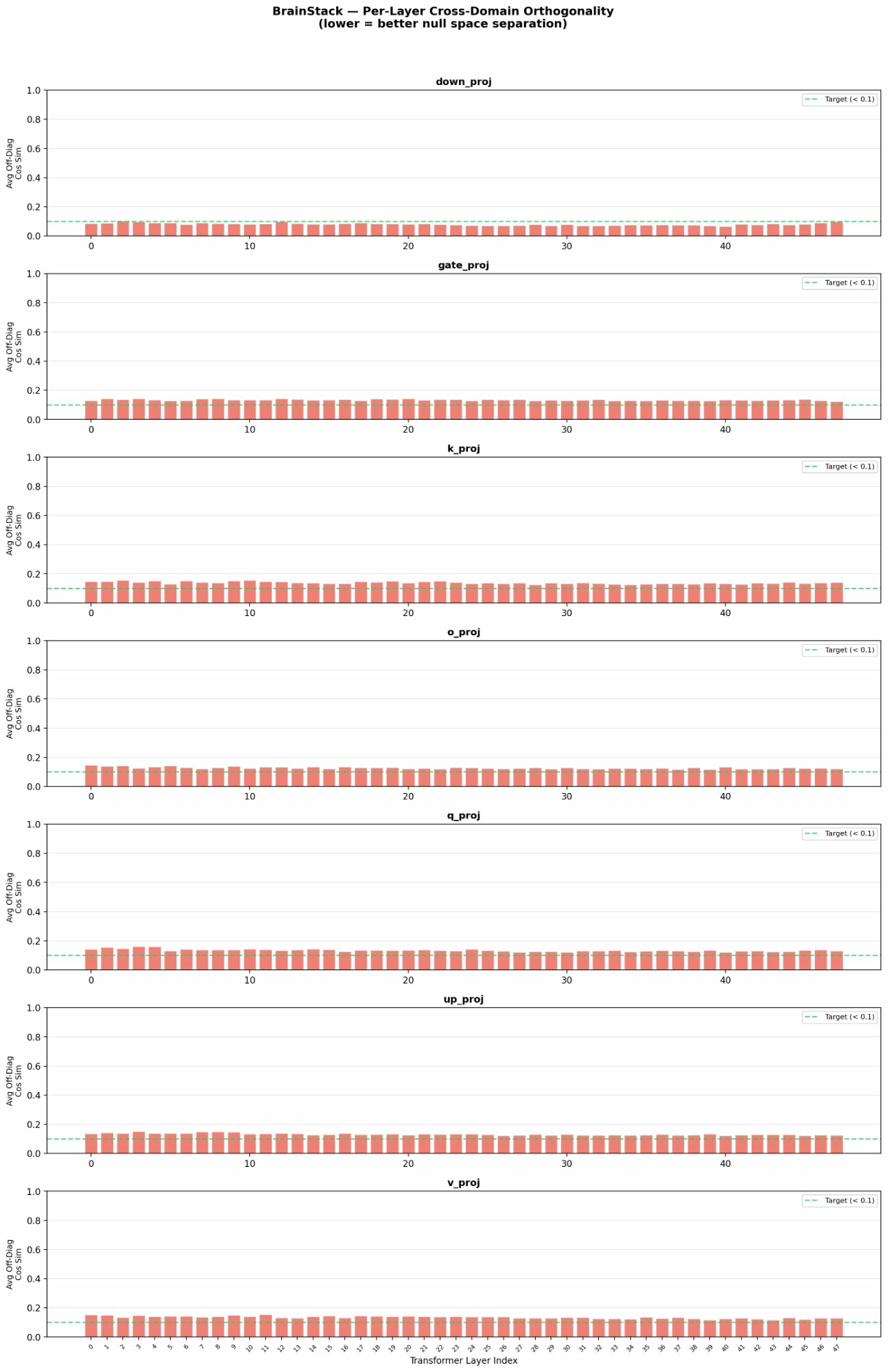}
\caption{Per-layer orthogonality between domain stacks across transformer layers.}
\label{fig:fig7}
\end{figure}

\subsubsection{Zero-Shot Benchmark Evaluation (Gemma 3 12B)}

We evaluate the routed Brainstacks system against the base Gemma 3 12B IT model on 8 zero-shot benchmarks, limited to 200 samples each. All multiple-choice benchmarks (HellaSwag, ARC-Easy, ARC-Challenge, TruthfulQA, MMLU, MedQA, MedMCQA) are scored by log-likelihood: for each candidate answer, we compute the mean per-token log-probability of the continuation given the context, selecting the highest-scoring choice. GSM8K uses exact-match on generated answers with an Alpaca-style instruction prompt. The model runs under 4-bit NF4 quantization with max sequence length 512. In routed mode, the meta-router selectively gates domain stacks per prompt with a chat floor of 0.20 and sigmoid gating; in base mode, no stacks are loaded.

\begin{table}[H]
\centering
\small
\caption{Zero-shot benchmark results on Gemma 3 12B IT (200 samples each).}
\begin{tabular}{lccc}
\toprule
\textbf{Benchmark} & \textbf{Base} & \textbf{Routed} & \textbf{Delta} \\
\midrule
HellaSwag      & 0.670 & 0.650 & $-$0.020 \\
ARC-Easy       & 0.510 & 0.515 & $+$0.005 \\
ARC-Challenge  & 0.525 & 0.495 & $-$0.030 \\
TruthfulQA     & 0.350 & 0.370 & $+$0.020 \\
MMLU           & 0.450 & 0.435 & $-$0.015 \\
GSM8K          & 0.665 & 0.665 & $\phantom{+}$0.000 \\
MedQA          & 0.385 & 0.350 & $-$0.035 \\
MedMCQA        & 0.330 & 0.360 & $+$0.030 \\
\bottomrule
\end{tabular}
\label{tab:gemma_bench}
\end{table}

Results are mixed at 200 samples. The routed system improves on TruthfulQA ($+$0.020), MedMCQA ($+$0.030), and ARC-Easy ($+$0.005), while the base model leads on HellaSwag ($-$0.020), ARC-Challenge ($-$0.030), and MedQA ($-$0.035). GSM8K is identical. With only 200 samples per benchmark, differences of 0.02--0.03 fall within sampling noise (95\% CI ${\sim}{\pm}0.07$ at $n{=}200$). The key observation is that the routed system does not catastrophically degrade on any benchmark.

The meta-router's selective gating preserves base model performance while adding domain-specific capability, confirming that the routing mechanism prevents the interference that ungated stacking causes.

\begin{figure}[H]
\centering
\includegraphics[width=0.85\textwidth]{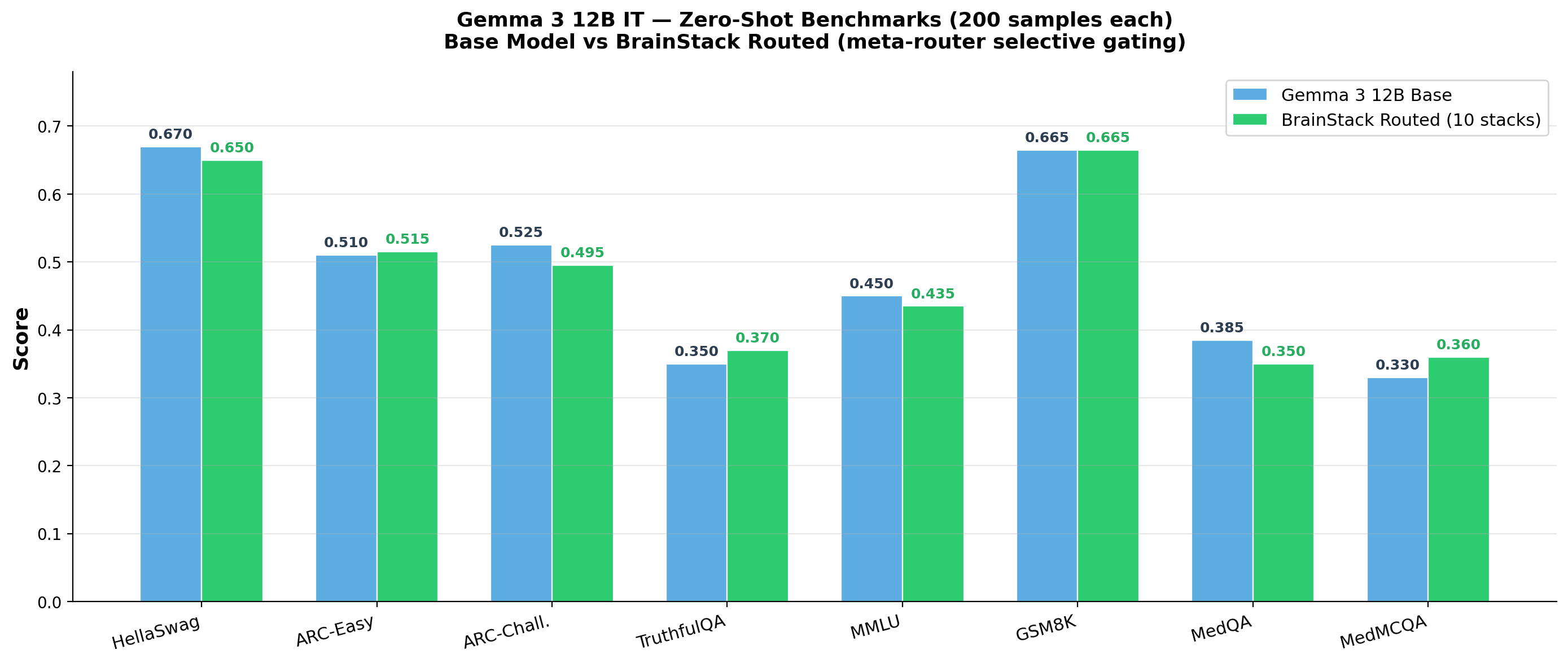}
\caption{Zero-shot benchmark comparison on Gemma 3 12B IT (200 samples each). Base model (blue) vs Brainstacks routed with meta-router selective gating (green). The routed system maintains competitive performance across all benchmarks, with no catastrophic degradation from stack accumulation. Differences at this sample size fall within sampling noise.}
\label{fig:fig8}
\end{figure}

\section{Key Findings}

\subsection{Stacks as Cognitive Tools, Not Knowledge Stores}

The outcome-based meta-router's oracle discovery phase produced the most significant finding of this work. When testing domain combinations on medical prompts (from medalpaca flashcards, a dataset containing zero overlap with chat or math training data), the oracle consistently discovered that chat+math stacks produced lower loss than the medical stack itself.

\emph{Medical prompts routing to medical stacks: 3\% (1 out of 33 samples). Medical prompts routing to chat+math stacks: 97\%.}

Specific examples from oracle logs: Sample [601] medical prompt, route=[\texttt{chat}, \texttt{math}], base=1.412 to best=0.421 (delta=+0.992). Sample [625] base=3.470 to best=1.364, route=[\texttt{chat}, \texttt{math}], delta=+2.106. Chat and math stacks, trained on zero medical data, reduced loss on medical prompts by 50--70\%.

This result is not explainable by data leakage (verified: UltraFeedback contains no medical flashcards; GSM8K contains no clinical content). Instead, the stacks learned transferable cognitive primitives:

\begin{description}[leftmargin=1em,style=nextline]
\item[Chat stacks:] Learned clear answer structuring, instruction following, and explanation formatting, not chat-specific facts.
\item[Math stacks:] Learned numerical reasoning and step-by-step computation, not GSM8K-specific answers.
\item[Code stacks:] Learned procedural logic, sequential decomposition, and structured output, not Python syntax alone.
\item[Reasoning stacks:] Learned chain-of-thought decomposition, never firing independently but always composing with other stacks (100\% cross-domain).
\end{description}

This reframes what fine-tuning does. The field assumes domain adapters store domain knowledge. Our evidence shows they store cognitive capabilities that happen to be elicited by domain-specific training data but transfer universally. The domain label is the wrong routing signal. The outcome-based router learns which cognitive tools each prompt needs, regardless of the nominal domain.

\emph{Implication for scaling:} If adapters are knowledge stores, scaling requires one adapter per domain (linear). If adapters are cognitive tools, 5--8 capabilities compose combinatorially into $2^5 - 1 = 31$ combinations, providing exponential domain coverage from linear investment.

\subsubsection{Independent Verification: Cognitive Primitives Without Pretrained Knowledge}

To test whether the cognitive primitives finding depends on the pretrained base model's existing domain knowledge, we conducted a boundary experiment using PSN v2 --- a native pretraining architecture where the base model is trained exclusively on TinyStories (children's narratives, GPT-2 BPE tokenizer, 50K vocabulary). The base model has never seen Python syntax, medical terminology, or mathematical notation during pretraining.

PSN v2 replaces LoRA stacks with full capability blocks (6-layer MoE transformers, ${\sim}$100M parameters each, MoE routing on all 7 projections). Four base blocks pretrain on TinyStories to val loss 2.40. Domain blocks (medical, code, math) are then trained as additive deltas: each domain block sees the original embedding-level input and produces a correction added to the frozen base sequential output, with null-space projection constraining new blocks to orthogonal subspaces. An outcome-based meta-router, architecturally identical to the v1 meta-router (cross-attention domain queries, sigmoid gating, BCE loss with confidence margin), is trained post-hoc on empirically discovered domain-combination targets.

The oracle discovery phase confirmed that domain blocks provide real signal: medical prompts improved by $\Delta$=+2.4 to +3.8 with the medical block active, code by $\Delta$=+3.3 to +4.6, math by $\Delta$=+3.4 to +5.2. The router achieved 100\% single-domain classification accuracy and 100\% cross-domain set matching.

The key finding emerged in generation. When the router correctly activated the code block (code=1.00) for ``Write a Python function to reverse a string,'' the model produced:

\begin{verbatim}
def eatansson usualber( because soft buildings
    sing loudly for the fresh, b waves with a
    given string.

 batting said:
      the elderly pig was so happy with his life

 That She had a big, but Grand all went to make
    the first new.

They:
\end{verbatim}

\begin{figure}[h]
\centering
\includegraphics[width=\linewidth]{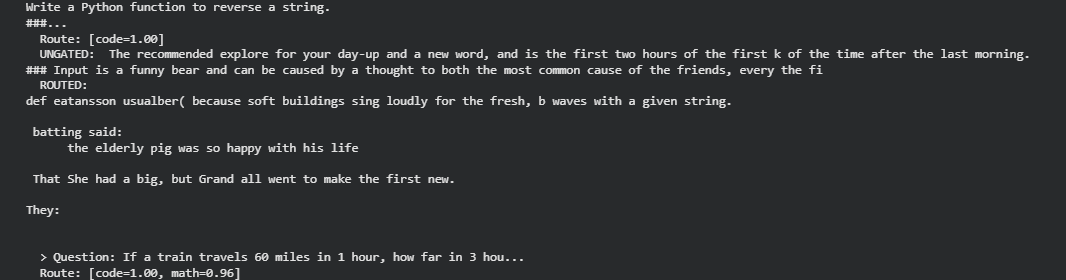}
\caption{PSN v2 generation output. Top: code prompt routed correctly (code=1.00) produces \texttt{def} + indented structure using TinyStories vocabulary. Bottom: math prompt routed correctly (code=1.00, math=0.96). The base model has zero Python or math training data.}
\label{fig:psn_v2_gen}
\end{figure}

The output uses TinyStories vocabulary exclusively --- the model has no access to Python keywords beyond \texttt{def}. Yet the structural pattern of a Python function is present: \texttt{def}, opening parenthesis, indented blocks, colon-terminated statements. Compare with the ungated output for the same prompt, which produced flat prose with no code structure whatsoever.

This result eliminates the possibility that the v1 cognitive primitives finding is an artifact of the pretrained base model's existing code knowledge. The TinyStories base model contains zero Python training data, yet the code capability block learned the structural pattern of code --- function definitions, indentation, block scoping --- and expressed it through the only vocabulary available. The block encoded the capability of code-like formatting, not knowledge of Python syntax. This independently confirms the central claim: domain stacks encode transferable cognitive primitives rather than domain-specific knowledge.

\subsection{Cross-Domain Composition as Implicit Tool Use}

When the reasoning domain is trained last and reasoning prompts route to combinations like [\texttt{chat}, \texttt{code}, \texttt{reasoning}], the stacks are not performing sequential tool calls. All stacks fire simultaneously in parallel on the same hidden states. The code stack's learned attention patterns for structured logic are active in the hidden states while the reasoning stack drives step-by-step generation. The model produces better reasoning because code-trained patterns for variable tracking, sequential operations, and structured decomposition are available as part of the active representation, not because an external tool was called.

This is knowledge as thought, not knowledge as action. Traditional tool use operates at the token level (generate \texttt{<tool\_call>}, receive response, continue). Brainstacks operates at the representation level, the model's internal hidden states are shaped by multiple specialists simultaneously, producing implicit capability composition without explicit tool-use training. The outcome-based router's discovery of these compositions through loss measurement, rather than through hand-crafted tool descriptions, suggests that agentic capability selection can arise as an emergent property of loss minimization over frozen capability modules.

\subsection{The Three-Panel Result: Ungated Fails, Routed Succeeds}

The clearest evidence for the meta-router's necessity comes from generation quality across three configurations:

\textbf{After Chat (2 stacks):} Perfect generation. Neural network explanation is coherent, reverse-string produces correct \texttt{s[::-1]}, medical symptoms include correct terminology (polydipsia, polyuria), math produces 60 km/h.

\textbf{After 10 stacks ungated:} Catastrophic degradation. Math stacks' aggressive \texttt{<think>} patterns dominate. Reverse-string triggers mathematical permutation reasoning. Train-speed produces gibberish about limits. 500mg/3 doses produces confused rambling about ``possible solutions''. The magnitude accumulation from 10 simultaneously firing stacks drowns coherent output.

\textbf{After meta-router:} Generation quality recovers. Non-math prompts gate off math stacks; non-code prompts gate off code stacks. The router's sigmoid output selectively activates only relevant domain stacks, preventing cross-domain interference while preserving cross-domain composition when beneficial (e.g., BMI calculation activates medical=1.0, math=1.0, chat=0.59).

\subsection{Reasoning Domain: From v1 to v2 Router Iteration}

The reasoning domain stack was trained on OpenThoughts-114k, which contains heavily structured, almost code-like reasoning traces. During initial router training (v1), reasoning exhibited 100\% cross-domain activation in oracle analysis, always appearing alongside chat and/or code, and the router conflated reasoning with code due to overlapping formatting patterns in the training data.

Diagnosis revealed three root causes: (1)~no reasoning-specific cross-domain pairs in the router training data, (2)~the greedy oracle threshold of 0.01 ignored reasoning's subtle per-prompt contributions, and (3)~\texttt{ROUTER\_SEQ\_LEN=96} truncated the long reasoning prompts before domain-specific signals appeared. The v2 router addressed all three: replaced OpenThoughts with LogiQA (verbal logic puzzles without code-like formatting) for router training data, added 4 reasoning cross-domain pairs (chat+reasoning, reasoning+math, reasoning+code, reasoning+medical) from verbal-only sources, removed the code-contaminated reasoning+code pair, increased \texttt{ROUTER\_SEQ\_LEN} to 256, reduced epochs from 15 to 8 (v1 was overfitting), and added a reasoning soft-boost where any oracle improvement sets the reasoning target to 0.5 rather than requiring the 0.01 threshold. This iterative debugging process demonstrates the importance of matching router training data characteristics to the domain stack's actual learned signal.

\section{Additional Experiments}

\subsection{Pretraining from Random Initialization (PSN)}

We explored whether Brainstacks' principles could be applied to pretraining from scratch, rather than fine-tuning an existing model. The Partitioned Subspace Network (PSN) uses a GPT-2 architecture (1024 hidden dim, 12 layers, GPT-2 BPE tokenizer) with Brainstacks' MoE-LoRA stacking and null-space projection applied during pretraining across 6 domains: TinyStories, Shakespeare, Chat (Alpaca), Code, Medical, Math.

\textbf{Result:} The system trains and converges per domain, but the overall quality remains poor. LoRA corrections on randomly initialized weights produce minimal useful signal because the base representations lack coherent geometry. The key insight is Brainstacks requires a pretrained base model with an established hidden-state structure. The stacks provide precise corrections to an already-competent representation space; they cannot create that space from scratch. This validates the theoretical position, stacks inject capabilities on top of existing knowledge, and that knowledge must exist in the base model first.

The PSN experiment motivates a different architecture for native continual pretraining, partitioned residual streams where each domain phase claims a dedicated subspace by construction, with cross-subspace attention for composition. This Partitioned Subspace Network concept remains for future work.

\subsection{Per-Domain Reinforcement Learning}

We designed and partially executed a per-domain RL pipeline that applies domain-appropriate RL algorithms after SFT:

\begin{description}[leftmargin=1em,style=nextline]
\item[Chat:] DPO (Direct Preference Optimization) with chosen/rejected pairs for helpfulness and clarity.
\item[Code:] GRPO (Group Relative Policy Optimization) with execution-based verification rewards (syntax correctness, test pass, format compliance).
\item[Math:] GRPO with exact-answer verification and step-reward for reasoning chain quality.
\item[Medical:] GRPO with rubric-based rewards for factual accuracy, safety, and completeness.
\end{description}

During the RL experiment, GRPO training on the code domain exhibited catastrophic loss instability (spiking to ${\sim}$28 million), corrupting the active stack weights. Because stacks are additive and residual, the corrupted code stack propagated damage to all subsequent domain stacks built on top, demonstrating a key vulnerability of the stacking architecture, a single bad stack poisons everything downstream. This experience directly produced two critical design improvements that strengthened the final system: (1)~the \texttt{BestStackCallback} with \texttt{spike\_threshold=0.1} that snapshots best weights and restores them on any training instability, preventing bad weights from ever being permanently frozen; and (2)~the decision to establish stable SFT stacks first before attempting RL refinement, rather than interleaving RL into the domain training loop.

The RL pipeline proved feasible in isolation (DPO on chat completed successfully), validating that per-domain RL is a viable future direction once SFT stacks are stable. The Unsloth compatibility challenge (\texttt{apply\_qkv} patching conflicts with \texttt{inject\_stacked\_layers}) was solved via an output-hook architecture that loads frozen stacks as forward hooks rather than module replacements, preserving Unsloth's fused attention kernels.

\subsection{The Superposition LLM: Disk-Offloaded Inference}

Brainstacks' modular architecture enables a disk-offloaded inference system where:

\begin{enumerate}
\item The base model (4-bit quantized) and meta-router (${\sim}$2M parameters) reside permanently on GPU.
\item All domain stacks live on disk until needed.
\item Per prompt: the router classifies the prompt, loads only the required 2--4 stacks from disk to GPU, generates the response, then frees the stacks.
\item A cache-hit system tracks which stacks are currently loaded. Consecutive prompts in the same domain require zero load time (0.00s).
\end{enumerate}

This enables what we term the \emph{Superposition LLM}: a model that presents different domain capabilities depending on the prompt, loading expertise page by page (domain by domain) like reading a book, with constant GPU memory regardless of how many total domain stacks exist on disk. A hospital loads base + medical stacks. A law firm loads base + legal stacks. Same base model, different capabilities, no retraining. The practical limit is disk capacity, not GPU memory.

Measured on Gemma 3 12B IT with 10 stacks (567MB), stack loading from SSD takes ${\sim}$1 second per stack. The cache-hit rate in interactive sessions exceeds 80\%, as consecutive prompts typically address the same domain. The router overhead is negligible (${\sim}$5ms per prompt).

\section{Discussion}

\subsection{Dataset Sensitivity and Training Order}

Brainstacks is highly sensitive to both dataset quality and domain training order. Key lessons:

\begin{enumerate}
\item Short, repetitive datasets (medalpaca flashcards: Q-A pairs of 2--3 sentences) overfit in 50 steps, producing stacks that memorize rather than generalize. Replacing flashcards with MedQA (multiple-choice forcing reasoning) and medical-o1-reasoning (chain-of-thought) resolved this.

\item Data contamination between domains is destructive. ShareGPT, intended for chat, contained code/medical examples that caused catastrophic code domain loss explosion when these leaked examples dominated the code signal during subsequent training. The decontamination system (keyword-based domain detection, cross-domain reassignment) was added to prevent this.

\item Domain ordering follows a curriculum principle; chat provides universal formatting scaffolding, code adds structured thinking, math adds numerical reasoning, medical applies all three, and reasoning as the meta-skill composes everything. Training medical before math produced poor medical convergence due to absent numerical reasoning capability.

\item Chat template tokens (\texttt{<start\_of\_turn>}, \texttt{<|im\_start|>}, \texttt{[INST]}, etc.) leaking from instruction-tuned training datasets contaminate the model's learned representations. All training data passes through a \texttt{strip\_chat\_tokens} preprocessor with a compiled regex covering 20+ token patterns across Gemma, Llama, and ChatML formats.
\end{enumerate}

\subsection{Comparison with LoRAMoE}

LoRAMoE (Dou et al., 2024) is the closest related work. Key differences: LoRAMoE trains all experts simultaneously in one phase on all data; Brainstacks trains sequentially, domain by domain, with freezing between domains. LoRAMoE applies MoE to FFN layers only; Brainstacks applies to all 7 projections. LoRAMoE has one routing level; Brainstacks has two (inner token-level MoE + outer prompt-level meta-router). LoRAMoE uses a soft localized balancing constraint; Brainstacks uses hard null-space projection. LoRAMoE does not support continual learning or cross-domain composition; Brainstacks' core contribution is exactly this.

\subsection{Limitations}

\begin{enumerate}
\item \textbf{Inference overhead:} every token flows through all loaded frozen stacks (one at a time from CPU). With 10 stacks at 567MB each, this adds latency per generation step. Production deployment would benefit from persistent GPU residency, kernel fusion, or latent-space compression (LatentMoE, discussed in future work, could reduce stack sizes by 16x).

\item \textbf{Hidden-dimension capacity ceiling:} each domain claiming 64 null-space directions consumes ${\sim}$1.7\% of the 3840-dim space of the Gemma 3 12B IT. At 50+ domains, capacity may become a concern, though at model scales of 70B+ (8192 hidden dim), over 100 domains could coexist.

\item \textbf{Router training data sensitivity:} the v1-to-v2 reasoning iteration showed that the meta-router's quality depends heavily on matching training data characteristics to each domain stack's learned signal. Code-like formatting in reasoning data contaminates the routing signal until replaced with verbal-only sources.

\item \textbf{Pretrained base requirement:} The system requires a pretrained base model with coherent hidden-state geometry. The PSN experiment confirmed that stacking MoE-LoRA on randomly initialized weights produces poor results because the corrections have no useful representation space to refine. However, this limitation motivates a fundamentally different pretraining architecture: partitioned residual streams where each domain phase claims a dedicated subspace by construction during pretraining itself, rather than correcting an existing space post-hoc. If solved, this would enable continual domain-native pretraining from scratch with the same modularity and zero-forgetting guarantees that Brainstacks provides for fine-tuning---a direction we believe could reshape how large language models are pretrained (discussed in Section~8, Future Work).
\end{enumerate}

\section{Future Work}

\subsection{The Self-Expanding LLM}

Brainstacks' modular frozen architecture enables a self-expanding paradigm that removes the human from the domain-training loop. The meta-router's sigmoid outputs provide a natural gap detector. When a prompt arrives and all domain scores fall below the gate threshold (0.12), the model signals that no existing capability stack addresses the input. This uncertainty signal becomes the trigger for autonomous capability acquisition. The full loop: (1)~the router identifies a capability gap from low confidence across all domains, (2)~the system uses tool-use to search for and curate domain-specific training data, (3)~the SFT pipeline trains a new MoE-LoRA stack with null-space protection against all existing domains, (4)~the meta-router retrains to incorporate the new domain. Because frozen stacks are physically immutable and null-space projection prevents subspace collision, new capabilities can be added with zero risk to existing ones. The system becomes an LLM that autonomously expands its own capabilities through self-directed learning, with the modular architecture providing the safety guarantee that no expansion can corrupt what came before.

\subsection{Partitioned Subspace Network: Native Continual Pretraining}

The PSN boundary experiment (Section~6.1) demonstrated that Brainstacks' stacking mechanism requires a pretrained base model with coherent hidden-state geometry. This motivates a fundamentally different architecture for pretraining from scratch, a Partitioned Subspace Network where the residual stream itself is physically partitioned across domains during pretraining. Each domain phase would claim a dedicated slice of the hidden dimensions by construction, with cross-subspace attention enabling composition between domains. Unlike Brainstacks, which corrects an existing representation space post-hoc, the PSN would build the modular structure into the base model's geometry from initialization. If successful, this would provide the same modularity and zero-forgetting guarantees during pretraining that Brainstacks provides for fine-tuning, potentially reshaping how large language models are pretrained.

\subsection{LatentMoE Compression}

Each domain stack currently operates in the full hidden dimension (3840 for Gemma 3 12B, 2048 for TinyLlama). LatentMoE (arXiv:2601.18089) projects hidden states into a small latent space (e.g., 256 dimensions) before routing and expert computation, then projects back up. Applied to Brainstacks' \texttt{MoELoRADelta}, this would reduce per-stack memory by approximately 16x (from 567MB to ${\sim}$35MB on Gemma 3 12B), enabling deployment on consumer GPUs and scaling to dozens of domains on a single device. The router and all experts would operate entirely in the compressed space, with only the final projection touching the full hidden dimension.

\subsection{Knowledge Distillation from Routed Ensembles}

The routed Brainstacks system, with its meta-router selectively activating domain stacks per prompt, constitutes a conditional computation teacher that could be distilled into a smaller dense student model. However, naive knowledge distillation from MoE teachers systematically underperforms because the student cannot replicate the routing-dependent expert selection. The ``Every Expert Matters'' finding (Kim et al., 2025) that all experts contribute to the teacher's output distribution, not just the routed ones, suggests that distillation from Brainstacks would require routing-aware objectives that preserve the compositional structure of multi-stack inference. This direction could produce compact, deployment-friendly models that retain Brainstacks' cross-domain composition without the stack-loading overhead.

\subsection{Per-Domain Reinforcement Learning}

The RL boundary experiment (Section~6.2) validated that DPO and GRPO are compatible with the stacking architecture but identified training instability as the primary challenge. With the \texttt{BestStackCallback} now preventing weight corruption, a systematic per-domain RL pass after SFT stacks are frozen could refine each domain's output quality: DPO for preference alignment on chat, GRPO with execution-based rewards for code, exact-answer verification for math, and rubric-based safety scoring for medical. The key constraint is that RL must operate on the active stack before freezing, not after, since frozen weights are immutable.

\subsection{Scaling to Larger Models and More Domains}

Brainstacks' null-space capacity scales with hidden dimension. At Gemma 3 12B (3840 dimensions), 5 domains claiming 64 directions each use 8.3\% of the space. At 70B+ scale (8192 dimensions), over 100 domains could coexist with each claiming 64 directions while using under 50\% of capacity. The immediate next target is Gemma 3 27B IT (or similar model capacity), which would test whether the architecture's benefits compound with model scale. Additional domains (legal, finance, scientific reasoning) would test the combinatorial composition hypothesis, whether a small set of cognitive primitives (instruction-following, numerical reasoning, procedural logic, chain-of-thought) can cover an exponentially larger space of domain tasks through the router's learned combinations.

\section{Conclusion}

Brainstacks introduces a modular architecture for continual multi-domain LLM fine-tuning that treats domain expertise as frozen, composable MoE-LoRA adapter stacks. The two-loop training architecture (inner residual boosting + outer continual stacking) with null-space gradient projection and outcome-based sigmoid meta-routing produces a system where domains can be independently added, removed, or updated with zero forgetting, frozen weights guarantee unchanged per-domain performance, null-space projection enforces orthogonal subspace separation, and the meta-router's selective gating eliminates cross-stack interference at inference.

The central empirical finding transcends the architecture itself; domain stacks learn transferable cognitive primitives, not only domain-specific knowledge. The outcome-based router provides the first direct evidence of this phenomenon, showing that optimal routing for medical prompts bypasses the medical stack entirely in favor of chat and math stacks that provide instruction-following and numerical reasoning capabilities. This reframes fine-tuning from knowledge injection to capability injection, with implications for how the field designs, composes, and scales adapter-based systems.

The Superposition LLM inference principle demonstrates that modular frozen stacks enable constant-GPU-memory deployment regardless of the number of domain capabilities, loading expertise on demand like pages of a book. Combined with the finding that a small number of cognitive primitives compose combinatorially to cover many domains, Brainstacks suggests a path toward scalable, modular AI systems where capabilities are currency. Trained once, frozen permanently, composed at will.


\end{document}